\documentclass[preprint,12pt]{elsarticle}

\usepackage{amssymb}
\usepackage{amsmath,amsfonts}
\usepackage{algorithmic}
\usepackage{algorithm}
\usepackage{array}
\usepackage[caption=false,font=normalsize,labelfont=sf,textfont=sf]{subfig}
\usepackage{textcomp}
\usepackage{stfloats}
\usepackage{url}
\usepackage{verbatim}
\usepackage{graphicx}
\usepackage{multirow}
\usepackage{tablefootnote}
\usepackage{multicol} 
\usepackage{arydshln}
\usepackage{xcolor}

\usepackage{times}
\usepackage{soul}
\usepackage{url}
\usepackage{booktabs}
\usepackage[utf8]{inputenc}
\usepackage[small]{caption}
\usepackage[hidelinks]{hyperref}
\usepackage{graphicx}
\journal{Neural Networks}

\begin{document}

\begin{frontmatter}

\title{Learning Constraints-Based Adaptive Hypergraph Neural Networks for Solving Vehicle Routing Problems}

\author[1]{Zhenwei Wang} 
\ead{zhenwei.wang@nottingham.edu.cn}
\author[2,4]{Tiehua Zhang}\ead{tiehuaz@tongji.edu.cn}
\author[1]{Jing Liu}
\ead{jing.liu@nottingham.edu.cn}
\author[1]{Heng Yu}
\ead{heng.yu@nottingham.edu.cn}
\author[3]{Kaizhu Huang}
\ead{kaizhu.huang@dukekunshan.edu.cn}
\author[1]{Ruibin Bai\corref{cor}}
\ead{ruibin.bai@nottingham.edu.cn}
\affiliation[1]{organization={University of Nottingham Ningbo China},
            addressline={199 East Taikang Road}, 
            city={Ningbo},
            postcode={315100}, 
            state={Zhejiang},
            country={China}}
\affiliation[2]{organization={Tongji University},
            addressline={1239,Siping Road}, 
            city={Shanghai},
            postcode={200092}, 
            state={Shanghai},
            country={China}}
\affiliation[3]{organization={Duke Kunshan University},
            addressline={8,Duke Road}, 
            city={Suzhou},
            postcode={215316}, 
            state={Jiangsu},
            country={China}}
\affiliation[4]{organization={State Key Laboratory for Novel Software Technology, Nanjing University},
            addressline={163,Xianlin Road}, 
            city={Nanjing},
            postcode={210023}, 
            state={Jiangsu},
            country={China}}
\cortext[cor]{corresponding author}

\begin{abstract}
The application of learning-based methods, particularly neural network based methods, to Vehicle Routing Problems (VRP) has emerged as a pivotal area of research within combinatorial optimization. VRPs are characterized by expansive solution spaces and complex constraints, often compounded by uncertainties, which render traditional approaches, such as exact mathematical models or heuristic methods, prone to substantial computational overhead. Although some recent learning-based methods have demonstrated good performance for VRPs with straightforward constraint scenarios, they often struggle to effectively handle the complex, hard constraints commonly encountered in practice. As the first work to incorporate hypergraph learning into routing
problems, this study introduces an end-to-end framework that integrates constraint-oriented hypergraph neural networks with reinforcement learning to address these challenges in vehicle routing problems. The central motivation is that routing constraints, such as capacity limits and time-window penalties, are naturally imposed on groups of nodes and partial routes rather than on isolated pairwise edges. Therefore, a representation mechanism capable of preserving such high-order constraint semantics is needed. A key innovation of this work lies in the development of a constraint-oriented dynamic hyperedge reconstruction strategy within the designated encoder, which substantially enhances hypergraph representation learning. Additionally, the decoder leverages a double-pointer attention mechanism to iteratively generate solutions. The proposed model is trained using asynchronous parameter updates guided by hypergraph constraints and optimized through a dual loss function that includes both constraint loss and policy gradient loss. Experimental results on benchmark datasets demonstrate that this approach not only eliminates the need for complex heuristic operators but also achieves state-of-the-art (SOTA) solution quality.
\end{abstract}


\begin{highlights}
\item To the best of our knowledge, this is the first application of hypergraph learning to vehicle routing problems for capturing higher-order node representations.
\item We develop an end-to-end hypergraph network with adaptive hyperedge construction that dynamically incorporates route constraints into the hyperedge set, enabling constraint-aware hypergraph representation learning within the encoder.
\item We propose a dual-pointer decoder that integrates global and local attention to balance exploration and exploitation during stepwise decision-making.
\end{highlights}


\end{frontmatter}



\section{Introduction}
\label{sec1}

The Vehicle Routing Problem (VRP) is a fundamental combinatorial optimization problem characterized by its inherent graph structure. It has garnered extensive research attention across logistics, operations research, and transportation. Due to its formulation as a discrete combinatorial decision problem, VRP is prone to combinatorial explosion when addressing large-scale instances, firmly establishing its status as a well-known NP-hard problem.

There are two primary approaches to solving such problems: exact and approximate methods~\cite{bai2023analytics}. Exact methods, such as linear programming and integer programming, leverage mathematical modeling to identify optimal solutions~\cite{xue2021hybrid}. However, due to the vastness of the solution space, these methods often become computationally intractable, especially for large-scale problems, as obtaining an optimal solution within a practical timeframe may be infeasible. To overcome this limitation, approximate methods are widely employed; common heuristics include local and neighborhood search~\cite{chen2020variable}, genetic algorithms~\cite{zhaoHybridGeneticSearch2024}, and genetic programming~\cite{chen2022cooperative}. These approaches start from a feasible solution and iteratively improve it via carefully designed operators to obtain near-optimal results. A major drawback, however, is their reliance on perfectly estimated problem parameters, often inducing over-optimization and poor robustness under uncertainties.

Recent advancements in artificial intelligence have renewed interest in machine learning-based approaches, offering robust and efficient approximation capabilities for combinatorial optimization. One line of work treats the VRP as a sequence-to-sequence decision problem, i.e., a constructive heuristic~\cite{liuHowGoodNeural2023a}. Inspired by natural language processing, these methods~\cite{nazariReinforcementLearningSolving2018,koolAttentionLearnSolve2019,kwonPOMOPolicyOptimization2020,XIAO2025107380,ZHU2023535,YANG2023589} adopt pointer networks~\cite{vinyalsPointerNetworks2017} and attention mechanisms~\cite{vaswaniAttentionAllYou2023} to generate solutions in an auto-regressive manner. Given the challenges associated with obtaining high-quality labeled solutions, unsupervised learning techniques, particularly Reinforcement Learning (RL), are frequently employed to train models on a large number of problem instances. By learning the underlying data distribution and node relationships, these approaches iteratively produce solutions that adhere to the required constraints. This end-to-end framework alleviates the need for extensive labeled datasets required in supervised learning, allowing the model to efficiently infer near-optimal solutions without relying on domain experts to design complex heuristic operators. Moreover, robustness, stability, delay effects, and synchronization have been extensively studied in competitive and memristive neural networks~\cite{chandrasekar2025resilient,subhashri2026robust}. These studies address control and synchronization issues and highlight the continuing importance of robust neural computation under dynamic constraints.

In contrast, alternative learning-based approaches, known as improvement heuristics, focus on optimizing feasible solutions rather than constructing them directly. They utilize machine learning to refine initial solutions, often augmented with perturbations such as 2-opt, 3-opt, or guided local search, to learn and improve the solution optimization process~\cite{dacostaLearning2optHeuristics2020}. Such an iterative improvement framework enables these methods to effectively balance exploration and exploitation, further enhancing solution quality. 

Regardless of whether a constructive or improvement approach is used, Graph Neural Networks (GNNs) have become increasingly popular due to the inherently graphical nature of the VRPs. However, most prior studies ~\cite{khalilLearningCombinatorialOptimization2017,maCombinatorialOptimizationGraph2019,joshi2019efficient,senuma2022gear,leiSolveRoutingProblems2022,wang2024gase} have primarily relied on pairwise-based GNNs, such as Random Walk Message Passing ~\cite{sugiyama2015halting}, Graph Convolutional Networks (GCN) ~\cite{hamiltonInductiveRepresentationLearning2017a}, and Graph Attention Networks (GAT) ~\cite{velickovicGraphAttentionNetworks2018a}, for node representation learning. A significant challenge in addressing the VRP lies in its limited feature space, which restricts the expressiveness and effectiveness of node representations. Additionally, routing-specific constraints are often underutilized during graph learning, creating performance bottlenecks. Moreover, these methods struggle to effectively capture high-order and sequential information. In step-by-step constructive approaches, where nodes are incrementally selected to form partial solutions, depending solely on low-order neighbor information can adversely affect the selection of subsequent nodes. This short-sightedness increases the risk of the overall solution becoming trapped in a local optimum, thereby limiting the quality of the final result. More importantly, capacity, time-window, route-duration, and precedence constraints are not determined by a single edge or by an isolated pair of customers. They are induced by a set of customers and by the partial route state accumulated before the current decision. Pairwise dense attention can connect all node pairs, but such connections do not preserve the identity of the constraint group; stacking more pairwise layers may also blur node representations and aggravate over-smoothing in deep message passing~\cite{chamberlain2021grand}. These limitations motivate a representation mechanism that treats constraint-induced customer groups as first-class structural objects rather than decomposing them into independent pairwise messages. Addressing these issues calls for the development of advanced methods capable of effectively utilizing problem-specific constraints and incorporating richer structural and sequential information into the learning process.

This paper proposes a novel end-to-end hypergraph learning framework for addressing the constraints in VRPs and their variants. The method enriches node representations by learning not only from low-order neighborhoods but also by dynamically integrating problem-specific constraint information into the hypergraph. Concretely, we construct constraint-oriented hyperedges to capture high-order neighbor relationships, which are then used to augment pairwise node representations. During solution generation, a multi-head attention mechanism iteratively produces approximate optimal solutions. The framework is trained with RL, enabling effective handling of multiple VRP constraints while achieving competitive performance. The novelty is therefore twofold. First, the encoder represents constraint-induced group relations through adaptive hyperedges instead of relying only on pairwise graph edges. Second, RL supplies route-level feedback to train the decoder without requiring optimal labels, while the hypergraph reconstruction loss guides the encoder toward constraint-aware representations. In this sense, reinforcement learning is the decision optimization mechanism, whereas hypergraph learning is the representation mechanism; their combination makes the framework both decision-aware and constraint-aware.

Key innovations of this work are summarized as follows:
\paragraph{Integration of Hypergraph Learning into Routing Problems} To the best of our knowledge, this is the first work to incorporate hypergraph learning into routing problems. A novel constraint-oriented dynamic hyperedge construction encoder is introduced, enabling the model to effectively capture high-order node relationships while satisfying problem-specific constraints.
\paragraph{Dual-Pointer Attention-Based Decoder with Route Recording} We propose a dual-pointer decoder, leveraging an attention mechanism that combines partial solution features and local node features. This design supports the auto-regressive generation of solutions while dynamically updating the hypergraph encoder using a reinforcement learning policy gradient, which ensures continuous improvement of the solution quality.
\paragraph{Extensive Experimental Validation} We have conducted extensive experiments to validate the effectiveness of our proposed model. The experimental results demonstrate that the hypergraph learning framework introduced in this paper achieves a leading level of performance improvement in VRP and its variants up to 7.38\%.\\

\section{Related Works}
This section provides an overview of learning-based methods for solving routing problems, categorized into graph-based and non-graph-based neural heuristics. Furthermore, as this work marks the first introduction of hypergraphs into routing optimization, advanced hypergraph-based methods will also be discussed.

\subsection{Non-graph Neural Heuristic}
The autoregressive solution generation approach was first introduced by~\cite{vinyalsPointerNetworks2017}, which proposed the Pointer Network (PtrNet). PtrNet utilized a long short-term memory network (LSTM) to encode node information and decode an output sequence, effectively addressing the challenge of mapping sequences with unequal lengths. However, its initial implementation relied on supervised learning, requiring large datasets with labeled optimal solutions. To address this limitation, Bello et al.~\cite{belloNeuralCombinatorialOptimization2017} extended PtrNet by adopting reinforcement learning, using node distances as rewards and training the model with policy gradient techniques, achieving impressive results. Nazari et al.~\cite{nazariReinforcementLearningSolving2018} further enhanced PtrNet by integrating LSTM with dot-product attention and applying reinforcement learning to tackle combinatorial optimization problems. Their approach outperformed Google's open-source solver, OR-Tools~\cite{ortools}, on the VRP. Inspired by the transformer architecture, Kool et al.~\cite{koolAttentionLearnSolve2019} introduced the attention-based encoder-decoder model AM, which recalculates node representations via attention in the encoding phase and outputs probability distributions over decision nodes in the decoding phase. Sampling from these distributions constructs the final solution. AM has since become a milestone in solving routing problems for non-graph structures.

Subsequent works have refined AM to improve performance, scalability, and generalization. For example, POMO~\cite{kwonPOMOPolicyOptimization2020} introduced feature transformations and multiple initial nodes to achieve near-optimal solutions for the Traveling Salesman Problem (TSP) and VRP. Bi et al.~\cite{bi2022learning} pre-trained AM using knowledge distillation to handle routing problems across data distributions and applied Lagrange relaxation to solve VRPs with temporal constraints, such as the Traveling Salesman Problem with Time Windows (TSPTW)~\cite{bi2024learning}. PointerFormer~\cite{jin2023pointerformer} and Tspformer~\cite{YANG2023589} extend the Transformer by, respectively, employing multiple attention-based decoders and sampled scaled dot-product attention to tackle large-scale TSP instances. Gao et al.~\cite{gao2023towardslocalglobal} enhanced AM's generalization by integrating local and global strategies, while Lin et al.~\cite{lin2024cross} improved its performance through pre-training and fine-tuning fully connected network parameters.

Additionally, hybrid approaches have combined learning-based methods with local search techniques. For instance, strategies like 2-opt are used to perturb feasible solutions, while heuristic operators are learned to further optimize them. These advancements have significantly improved the efficiency, scalability, and generalization of learning-based methods for routing problems.

More recently, the field has moved beyond single-task synthetic benchmarks toward broader generalization. Omni-VRP studies fast adaptation across problem sizes and distributions through a meta-learning framework~\cite{zhou2023omni}. RouteFinder formulates multiple VRP variants as attribute combinations in a unified environment and investigates foundation-model-style pretraining for routing~\cite{berto2024routefinder}. RL4CO provides a modular benchmark and implementation framework for reproducible RL-based combinatorial optimization research~\cite{berto2025rl4co}. Heavy-decoder methods such as LEHD shift computational effort from a static encoder to a dynamic decoder and show strong cross-size generalization on TSP and CVRP~\cite{luo2023lehd}. The light-decoder method further analyzes how decoder context and multi-expert routing policies affect VRP generalization under different constraint regimes~\cite{huang2025rethinking}. For complex constraints, Bi et al.~\cite{bi2024learning} propose proactive infeasibility prevention, showing that local feasibility masks may be insufficient when future infeasibility is induced by the current action. The more recent CaR framework explicitly studies construct-and-refine constraint handling, indicating that representation sharing and feasibility refinement are becoming central issues for neural routing solvers~\cite{bi2026car}. Scale-Net further studies cross-scale and multi-task routing through a hierarchical U-Net structure~\cite{liu2026scalenet}, while EvoReal addresses the gap between synthetic and real routing distributions through LLM-guided instance generation and progressive adaptation~\cite{zhu2026evoreal}. These studies clarify the current state of the art: neural routing is progressing toward unified, scalable, constraint-aware, and distribution-robust solvers. In contrast, the present paper focuses on a complementary direction: introducing hypergraph learning to explicitly encode high-order constraint relations within the routing representation itself.

\subsection{Graph-Based Neural Heuristic}

While prior works have explored sequence-to-sequence methods for solving routing problems, they largely overlook the underlying topological structure of these problems. To address this limitation, GNN-based approaches have gained prominence. Khalil et al.~\cite{khalilLearningCombinatorialOptimization2017} employed struct2vec to represent routing problems as graphs and trained the model using a deep Q-network. Joshi et al.~\cite{joshi2019efficient} proposed a non-autoregressive framework for the Traveling Salesman Problem (TSP), leveraging stacked GCN layers to generate an edge adjacency probability matrix. This matrix was visualized as a heat map, and solutions were extracted via beam search. Ma et al.~\cite{maCombinatorialOptimizationGraph2019} introduced a hierarchical graph pointer network with a graph embedding layer to capture node relationships. Their reinforcement learning-based approach employed layer-specific reward functions to stabilize training and improve generalization for the TSP. Similarly, Senuma et al.~\cite{senuma2022gear} used GCN to extract node and edge features, which were integrated with a pointer network to construct a policy model. By minimizing total edge costs through reinforcement learning, they proposed a re-planning strategy to optimize paths. Lei et al.~\cite{leiSolveRoutingProblems2022} further advanced GNN-based methods by utilizing GAT for graph representation learning, incorporating edge features into the encoder outputs. Their end-to-end framework, trained via reinforcement learning with policy gradients, achieved high-quality solutions for both TSP and VRP. Wang et al.~\cite{wang2024gase} improved graph representation by designing a graph attention encoder that filtered globally low-correlation nodes via sampling, thereby enhancing node-level representations.

Despite these advancements, most GNN-based approaches focus on pairwise relationships between nodes and edges, neglecting higher-order interactions within graph-structured data. This limitation is particularly pronounced in vehicle routing problems, where constraints such as vehicle capacity often involve group interactions among nodes rather than simple pairwise relationships. For example, fulfilling capacity constraints may require coordination among multiple nodes, including those without direct connections. Thus, methods relying solely on pairwise relationships may struggle to address the complexity of routing problems with diverse constraints. This observation is consistent with the broader graph-learning literature: pairwise GNNs are effective for local relational propagation, but deeper propagation can make node embeddings less distinguishable, especially when all customers repeatedly exchange similar messages in dense routing graphs~\cite{chamberlain2021grand}. A hyperedge offers a different inductive bias because the aggregation unit is a constraint-related node set rather than a single pairwise edge.
\subsection{Hypergraph Learning}
Hypergraph learning extends traditional graph learning by modeling high-order correlations among data through hyperedges that connect multiple nodes, enabling the representation of complex structures in high-dimensional and nonlinear spaces~\cite{zhang2025learning}. Several notable studies have advanced this field. Guo et al.~\cite{guo2021hierarchical} employed hypergraphs to model group similarity in recommendation systems, framing group preference learning as an inductive hyperedge embedding task. Their approach connects all groups as a hypergraph to achieve accurate group representations. Gao et al.~\cite{gao2022hgnn+} proposed HGNN+, a general framework for hypergraph neural networks that captures high-order relationships by linking multimodal and multi-type data through hyperedges and hyperedge groups, effectively modeling intricate data dependencies. Jiang et al.~\cite{jiang2019dynamic} introduced the dynamic hypergraph neural network (DHGNN) framework, which consists of two key components: dynamic hypergraph construction (DHG) and hypergraph convolution (HGC). The DHG module uses k-NN to generate initial hyperedges and extends them with k-means clustering, capturing both local and global relationships. The HGC component encodes high-order data correlations within the hypergraph structure, enhancing the model's capacity to represent complex dependencies. The above studies emphasize that compressing high-order events into pairwise edges can discard the semantics of the group itself. For routing, this point is particularly relevant because the feasibility of a partial route is defined by the joint demand, arrival time, and service state of multiple customers.

Building on these advancements, this paper extends hypergraph-based techniques to address the VRP. In this approach, hyperedges are dynamically constructed according to problem-specific constraints and high-dimensional projections of the nodes. The model is trained using feedback from deep reinforcement learning, enabling it to effectively capture and optimize the high-order relationships inherent in VRP.

\section{Preliminaries}
This section begins by introducing the graph-based modeling approach for the routing problem, including the definition of hyperedges within the hypergraph structure. Subsequently, the problem is abstracted into a Markov Decision Process (MDP)~\cite{karimiLevelReasoningDeep2023} and addressed using the designated reinforcement learning approaches.
\subsection{Problem Formulation}
The VRP can be formulated on a graph \(\mathcal{G}=(\mathcal{V},\mathcal{A})\), where \(\mathcal{V}\) is the node set including a depot node (node 0) and a set of customer nodes to be serviced. \(\mathcal{A}\) is the set of arcs between nodes. Vehicles depart from the depot, service the demand of each customer node exactly once, and then return to the depot. Each vehicle is subject to a capacity constraint, ensuring that the total demand of the customer nodes it serves does not exceed its maximum capacity. Formally, it is called the Capacitated Vehicle Routing Problem (CVRP). The objective is to determine the shortest route while satisfying all given constraints. The mathematical model of CVRP in set partitioning form can be expressed as follows:

    \begin{equation}\label{eq1}
    \min \sum_{r \in \Omega} c_r z_r
\end{equation}
subject to
\begin{equation}\label{eq2}
    \sum_{r \in \Omega} z_{r}=|\mathcal{K}|
\end{equation}
\begin{equation}\label{eq3}
    \sum_{r \in \Omega} a_{i,r} z_r=1 ,\quad \forall i \in \mathcal{V} \backslash\{0\}
\end{equation}
\begin{equation}\label{eq4}
    z_r ,a_{i,r} \in \{0,1\},\quad \forall r \in \Omega,\forall i \in \mathcal{V} \backslash\{0\}
\end{equation}

Expression (1) defines the objective function of the CVRP in its set partitioning form. Here, \(\Omega\) denotes the set of all feasible routes satisfying vehicle capacity constraints, \(c_r\) represents the cost of route \(r\in \Omega\), and \(z_r\) is a binary decision variable indicating whether route \(r\) is included in the final solution (\(z_r = 1\)) or not (\(z_r = 0\)). The objective is to minimize the total cost by selecting a set of feasible routes that collectively form a feasible solution. Constraint (2) restricts the number of selected routes to the number of available vehicles \(\mathcal{K}\). \(a_{i,r}\) is an indicator that takes a value of 1 if route \(r\) covers node \(i\), and 0 otherwise. Constraint (3) ensures that each node \(i\) is visited exactly once across all vehicles, while Constraint (4) enforces the binary nature of the decision variables.

\paragraph{Pairwise Graph Definition}Consider a pairwise graph \( \mathcal{G} = (\mathcal{V}, \mathcal{E}) \), with node set \( \mathcal{V}=\{v_0,v_1,\ldots,v_n\} \), where \( v_0 \) represents the warehouse, while \( v_1 \) to \( v_n \) denote the customer nodes. The edge set \( \mathcal{E} = \{(v_i,v_j)\mid v_i,v_j \in \mathcal{V}, i \neq j \} \) represents the connections between nodes, indicating the potential paths that vehicles may travel.

\paragraph{Hypergraph Definition} Given a hypergraph \( \mathcal{\hat{G}} = (\mathcal{V}, \mathcal{\hat{E}}) \), where \( \mathcal{V} = \{v_0, v_1, \dots, v_n\} \) is the node set and \( \mathcal{\hat{E}} = \{e_1, \dots, e_m\} \) is the hyperedge set, each hyperedge \(e_k \in \mathcal{\hat{E}}\) is defined as a subset of nodes, $e_k = \{v | v \in \mathcal{V}\},|e_k|=\Delta$. The degree of a hyperedge, \(\Delta\), represents the number of nodes it contains. Hyperedge formation exhibits strong aggregation properties, grouping highly correlated nodes under heterogeneous constraints. If each hyperedge \(e_k\) satisfies Constraint (3) and forms subsets of \(\Omega\), i.e., \(e_k \subseteq r\in\Omega\), this structure effectively supports the sequential decision-making process. A useful way to understand this advantage is to compare a hyperedge with clique expansion. A hyperedge of degree \(\Delta\) can be converted into pairwise edges by connecting all node pairs, but this introduces \(O(\Delta^2)\) pairwise relations and no longer explicitly preserves the fact that these nodes belong to the same constraint-induced group. Direct hyperedge aggregation keeps the group identity and allows the encoder to associate the node set with a route-level constraint. This is important for CVRP and CVRPTW, where feasibility depends on the cumulative demand or temporal status of a set of visited and candidate nodes rather than a single pairwise distance.

\subsection{Reinforcement Learning}
This paper formulates the CVRP as a sequential decision-making optimization problem in discrete space. Starting from the initial city layout of a specific problem instance, it incrementally constructs a city arrangement to guide decision-making. This process is modeled as a Markov Decision Process (MDP), where the states, actions, and rewards are defined and processed within the reinforcement learning framework.

\paragraph{State}The state of the CVRP consists of two key components: the dynamic state, which includes the current locations of nodes to be visited and the remaining vehicle capacity, and the static state, representing the fixed locations of all nodes. State transitions are determined by the selected node based on the action taken. As each node is selected, the vehicle's capacity is updated. When the vehicle returns to the depot, its capacity is fully reset.
\paragraph{Action}
The action corresponds to the decision-making process of selecting the next node. During the decoding stage, the system evaluates the current state and samples a node from the set of accessible nodes, based on a probability distribution generated by the attention mechanism of the decoder network. The conditional probability of selecting a node at step \(t\) through policy \(\pi\) is denoted as \(p = (\pi_t \mid \pi_0,\ldots,\pi_{t-1}, s_t)\), where \(s_t\) is the state of the current step. With the influence of neural network parameters \(\theta\), the solution probability of an instance \(\mathcal{G}\) can be represented as: 
\begin{equation}\label{eq5}
    p_{\pi_{\theta}}(\tau|\mathcal{G}) = \prod_{t=1}^{|\tau|} p_{\pi_{\theta}} (\tau_t \mid \mathcal{G}, \tau_{t^\prime}, \forall t^\prime<t )
\end{equation}
In this context, \(\tau\) denotes a trajectory derived from policy \(\pi\), influenced by the neural network parameters \(\theta\).

\paragraph{Reward}
The objective of CVRP is to determine the shortest path. Meanwhile, in reinforcement learning, the goal is typically to identify a policy that maximizes rewards. To align this with the CVRP, we use the negative value of the route distance obtained for each instance as the reward for the solution generated by the policy. Therefore, the reward for the solution \(\tau\) can be expressed as follows:
\begin{equation}\label{rewardcal}
    r(\tau) = -L(\tau) = -\sum_{i=0}^{|\tau|}\left\|\tau_{i}-\tau_{i+1}\right\|_2
\end{equation}

Overall, reinforcement learning employs Monte Carlo sampling~\cite{zheng2021combining} to estimate rewards for various solutions under a specific policy. It modifies neural network parameters through gradient ascent to maximize rewards, that is, minimize the total route distance.
\section{Methodology}
\begin{figure*}[thb]
\centering
\includegraphics[width=1.0\textwidth]{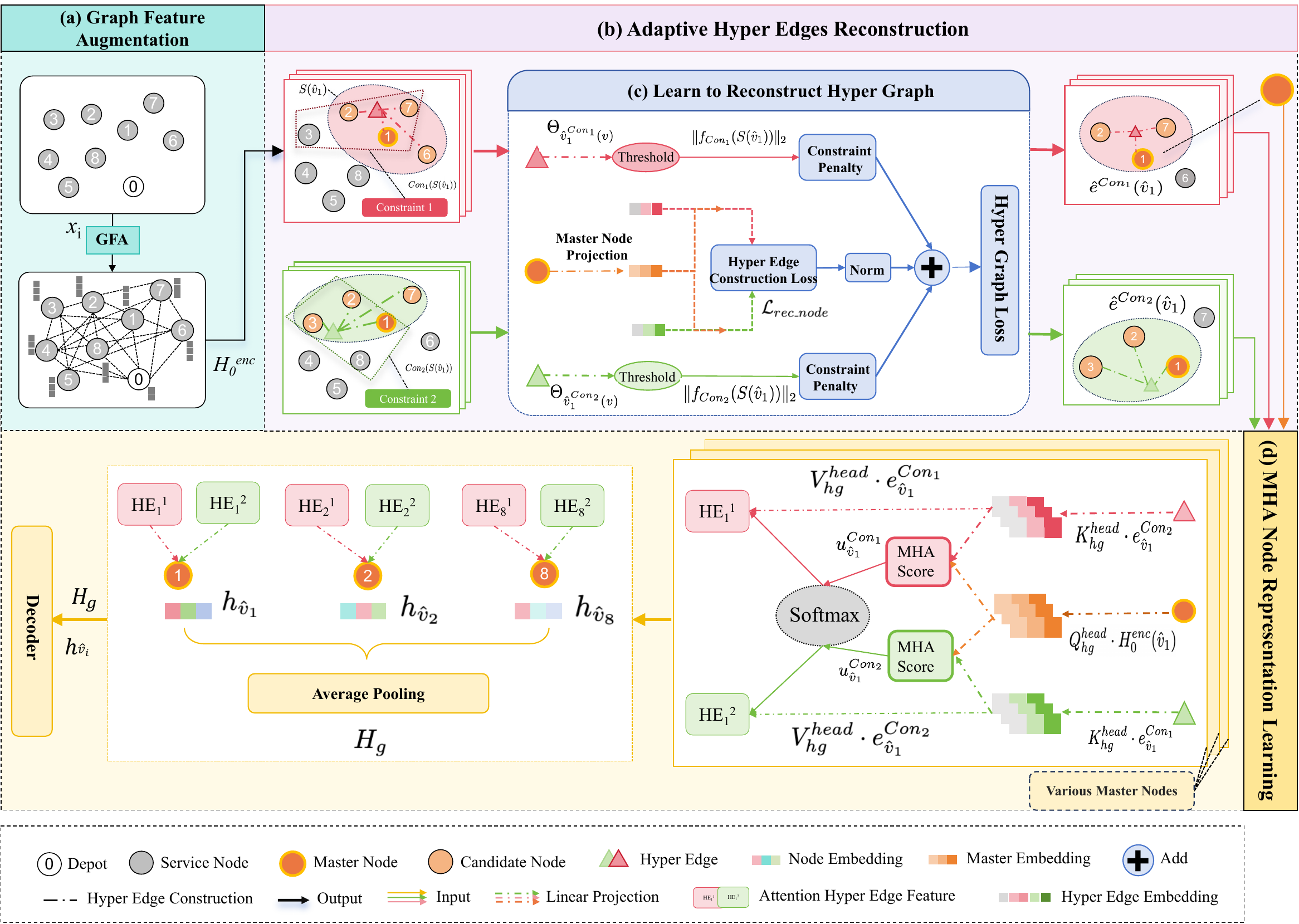}
\caption{Encoding Process Based on Adaptive Hypergraph Learning. In module (a), the original node coordinates are enhanced to generate features for the encoder. Module (b) maps nodes through a linear layer to identify hyperedge candidates based on a threshold. In module (c), candidates are validated against constraints, and a loss is computed to optimize node correlation within the hyperedge. Finally, module (d) uses multi-head attention to fuse hyperedge and node representations, averaging the output as the graph representation for the decoder.
}
\label{figure1}
\end{figure*}
The proposed constraint-oriented hypergraph learning framework provides a comprehensive end-to-end learning solution in an Encoder-Decoder manner. First, the encoder dynamically constructs hyperedges for each node while ensuring compliance with various constraints. A multi-head attention mechanism is then applied to integrate hyperedge representations with original node features, enabling effective graph representation learning. The resulting node representations are subsequently passed as input to the decoder. The decoder operates autoregressively, leveraging a dual attention pointer mechanism that incorporates both the current node and historical nodes as context embeddings to compute local and partial solution attention scores. Finally, nodes are sequentially generated based on the learned probability distributions. The following subsections provide a detailed explanation of the key components of this architecture.

\subsection{Graph Feature Augmentation}
In light of previous works, the features of nodes in routing problems are relatively sparse (usually the coordinates or distances of nodes in Euclidean space), providing limited useful information. To address this issue, we perform data augmentation at the node level. Drawing inspiration from POMO \cite{kwonPOMOPolicyOptimization2020}, we apply transformations such as rotation, symmetry, and folding to represent diverse spatial states of the same instance. Unlike POMO, however, we do not increase the number of instances. Instead, the augmented results are incorporated as extended features for each node, facilitating the averaging process in the subsequent attention mechanism. 
\begin{table}[!ht]
\caption{Data Augmentation}
\label{tab:dataaug}
    \centering
    \begin{tabular}{lccc}
    \toprule
      Original Instance&\multicolumn{3}{c}{($x,y)$}  \\
      \midrule
         Augmented Instance&$f(y,x)$& $f(1-x,1-y)$ & $f(\rho,\alpha)$\\
            \bottomrule
    \end{tabular}
\end{table}
Operations such as rotation and folding on coordinates are essentially permutations and combinations of the original coordinates \((x, y)\) and flipped coordinates \((1-x, 1-y)\). Instead of generating different forms of the same instance, we integrate these as new features, reducing POMO's eight coordinate mappings to two owing to the permutation-invariant nature of graph structures. Following the setup in~\cite{gao2023towardslocalglobal}, we further incorporate polar coordinates representing the distance and direction of nodes relative to the depot, where \(\rho\) is the radial distance and \(\alpha\) is the angular direction. This introduces directional information between nodes and the depot into the features. Through this data augmentation, we aim to provide the neural network with richer information about symmetry and directionality, enabling it to consider more than just the 2D planar coordinates of nodes during decision-making.
Additionally, we introduce polar coordinates as enhanced features, encoding both the distance and angle between nodes. These three types of coordinate mapping relationships are shown in Table \ref{tab:dataaug}. After applying these mappings to the coordinate values, they are used as new feature dimensions for concatenation. Finally, all augmented features are structured into a graph format, where a graph attention network layer is employed to fuse pairwise node representations effectively. The augmented node embeddings can be calculated as follows:

\begin{equation}\label{eq6}
    \boldsymbol{x}_i =
       \boldsymbol{f}_{aug}(\boldsymbol{v}_i)||\boldsymbol{f}_{pola}(\boldsymbol{v}_i)\quad i = 0,\ldots,n
\end{equation}
\begin{equation}\label{eq7}
    \boldsymbol{H}_0^{enc} = \left\{
    \begin{array}{ll}
       GAT(BN(FF(\boldsymbol{x}_i)))  & i = 0 \\ 
       GAT(BN(FF(\boldsymbol{x}_i||\boldsymbol{q}_i)))  &  i = 1,\ldots,n\\
    \end{array}
    \right.
\end{equation}

The two mapping functions in Equation (\ref{eq6}) enhance the original node coordinates \(v_i\) in two ways: \(\boldsymbol{f}_{aug}(\cdot)\) applies transformations such as flipping, rotating, and folding to the Euclidean coordinates in two-dimensional space, while \(\boldsymbol{f}_{pola}(\cdot)\) converts the coordinates into polar form and \(\cdot||\cdot\) denotes concatenation. Equation (\ref{eq7}) further captures the pairwise representations of nodes based on the enhanced attributes \(\boldsymbol{x}_i\). It is important to note that depot and service nodes are treated separately: the demand feature of a service node is represented by \(q_i\), whereas the depot \(v_0\) has no service demand. Here, \(FF(\cdot)\) refers to a feedforward neural network layer, \(BN(\cdot)\) denotes batch normalization~\cite{ioffe2015batch}, and \(GAT(\cdot)\) represents a graph attention network layer. The use of $GAT$ encoding allows the model to learn pairwise features prior to hypergraph learning, ensuring that low-order pairwise information is preserved when dynamically constructing hyperedges. The input for hypergraph representation learning in the encoder is denoted as \(H_0^{enc}\).

\subsection{Encoder with Hypergraph Learning}
The proposed hypergraph learning-based encoder is composed of two key modules: \textit{Constraint-oriented Adaptive Hyperedge Reconstruction} and \textit{Attention-based Node Representation Learning}. In the first module, hyperedges are dynamically constructed to adaptively learn the weight coefficients of nodes during training. A filtering process then determines which nodes will compose each hyperedge. In this way, the composition of hyperedges for each node includes highly correlated nodes that are limited by specific constraints. Once the hyperedges are constructed, their features are updated at the node level using a multi-head attention mechanism in the second module. This process integrates the information from hyperedges with the original node representations, ultimately generating the output graph representation and completing the encoding process. The entire encoding procedure is illustrated in Fig. \ref{figure1}. Detailed explanations of these two core modules are provided in the following subsections.

\subsubsection{Selection of Hyperedge Nodes}
As illustrated in Fig. \ref{figure1}(b), a master node \(\hat{v}_i\) forms its associated hyperedge node set \(\hat{e}(\hat{v}_i)\) by selecting nodes around itself. All nodes, including the master node, are first projected into a high-dimensional space through a linear transformation layer \(\Theta\). Feature similarity is then measured via a dot product, and nodes exceeding the threshold \(\delta\) are selected as candidates, ensuring strong feature-space correlation within the hyperedge. Additionally, the candidate set \(\mathcal{S}(\hat{v}_i)\) must satisfy problem-specific constraints. To handle the complexity of routing problems, heterogeneous hyperedges are constructed for each master node \(\hat{v}_i\), with each hyperedge tailored to a specific constraint \(Con_j\). This approach ensures that all nodes within a hyperedge adhere to the corresponding constraint. To capture the full problem context, each node takes turns as a master node, enabling the construction of diverse constraint-based hyperedges.

The selection of the hyperedge node set \(\hat{e}(\hat{v}_i^{Con_j})\) 
based on the master node \(\hat{v}_i\) is illustrated in Equations (\ref{candinode}) and  (\ref{hedgenode}).
\begin{equation}\label{candinode}
    \mathcal{S}(\hat{v}_i) = \{v | v \in \mathcal{V}\backslash\{\hat{v}_i\}, \Theta_{\hat{v}_i}(v) > \delta \}
\end{equation}
\begin{equation}\label{hedgenode}
    \hat{e}^{Con_j}(\hat{v}_i) = \{v | v \in \mathcal{\hat{S}}(
    \hat{v}_i),\mathcal{\hat{S}}(\hat{v}_i) = Con_j(\mathcal{S}(\hat{v}_i)) \}
\end{equation}

In Equation (\ref{candinode}), \(\Theta\) denotes the weights of the linear layer associated with each instance node, with \(\hat{v}_i\) as the master node. The threshold hyperparameter \(\delta\), whose sensitivity is analyzed in the next section, acts as a filtering mechanism. It maps the weights to the master node \(\hat{v}_i\) and all other nodes \(v \in \mathcal{V}\backslash\{\hat{v}_i\}\), represented as \(\Theta_{\hat{v}_i}(v)\).

In Equation (\ref{hedgenode}), \(Con_j\) captures the influence of the \(j\)th constraint on the candidate node set \(\mathcal{S}(\hat{v}_i)\). A penalty-based mechanism is applied, allowing nodes that violate constraints to incur penalties rather than being strictly excluded. This soft constraint approach provides flexibility, especially in routing problems where constraints like time or cost penalties are critical. For example, visiting a node that violates a constraint results in a penalty instead of outright exclusion, enabling adaptable solutions for problem variants. Here, we denote \(\mathcal{\hat{S}}(\hat{v}_i)\) as the candidate node set for hyperedges after the application of constraint penalty feedback. This penalty feedback is essentially a mean squared error (MSE) applied to various constraint values, such as capacity limitations. We will provide further clarification on the calculation of the hyperedge constraint loss in subsequent sections.
For strict constraint conditions, masks are applied in the decoder to enforce compliance. This dual approach enables the effective handling of both soft and hard constraints. 
Ultimately, the various constraint-based hyperedges associated with the master node \(\hat{v}_i\) (\(\forall \hat{v}_i \in \mathcal{V}\)) can be expressed as:
\( \mathcal{\hat{E}}(\hat{v}_i) = \hat{e}^{Con_j}(\hat{v}_i), \quad i=\{1, \ldots, n\} , \, j=\{1, \ldots, m\}
\). This formulation ensures that heterogeneous hyperedges are constructed to address diverse constraint types, enabling the model to adapt to the specific requirements of different routing problem variants.

The above formulation also provides a theoretical justification for using hypergraphs in routing. In a pairwise GNN, the message passing unit is an edge and therefore describes a relation between two nodes. However, capacity feasibility is determined by the accumulated demand of a subset of customers, and time-window feasibility depends on the temporal state induced by a partial route. These constraints are set functions over multiple nodes. By associating each master node with constraint-specific hyperedges, the encoder approximates these set functions through high-order aggregation instead of decomposing them into independent pairwise messages. This does not constitute an optimality guarantee, but it provides a structurally more faithful inductive bias for VRPs with group-level constraints.

\paragraph{Learning to Reconstruct Hyperedge Nodes}
As shown in Fig.\ref{figure1}(c), hyperedge nodes are dynamically reconstructed during training based on the neural network's weight mapping \(\Theta\). To enable adaptive parameter tuning, we propose a composite loss function with two components: node correlation loss and hyperedge constraint loss.

The node correlation loss measures the feature similarity between the master node and its hyperedge nodes. Minimizing this loss ensures semantic consistency and improves the representational quality of the hyperedge, resulting in a more cohesive and meaningful hypergraph structure. The hyperedge constraint loss enforces problem-specific constraints by penalizing violations within the hyperedge. Instead of excluding non-compliant nodes, penalties are applied based on the degree of violation, allowing flexible handling of soft constraints. This design ensures adaptability to diverse scenarios while maintaining feasibility under hard constraints, enhancing the model's robustness and applicability to complex tasks.

The computation of node correlation loss is as follows:
    \begin{equation}\label{nodeloss}
    \resizebox{.91\linewidth}{!}{$
            \displaystyle
    \mathcal{L}_{node} = \sum_{i=1}^n\sum_{j=1}^m\left\| H_0^{enc}(\hat{v}_i)\cdot \theta_{proj} - \Theta_{\hat{v}_i}^{Con_j}(v)\cdot H_0^{enc}(\hat{e}^{Con_j}(\hat{v}_i))   \right\|_2
    $}
\end{equation}
\(H_0^{enc}(\cdot)\) represents the augmented node representation, while \(\theta_{{proj}}\) denotes the weights of a linear layer that projects the master node into a high-dimensional space. Equation (\ref{nodeloss}) computes the MSE between the master node and its associated slave nodes within hyperedges, considering various \(m\) constraints in the high-dimensional space. A lower MSE reflects a stronger feature correlation among hyperedge nodes.

The reconstruction of hyperedge nodes is guided by the learnable reconstruction projection coefficient vector \(\Theta_{\hat{v}_i}^{Con_j}(v)\) for constraints, where \(\hat{v}_i\) represents the master node and \(v \in  \mathcal{V}\backslash\{\hat{v}_i\}\),\(j={1,...,m}\). We use two regularization terms in this coefficient vector to balance sparsity and smoothness when dynamically reconstructing hyperedges, as described in Equation (\ref{recnodeloss}), where the subscripts of \(\hat{v}\)  and \( Con \) are omitted to represent them in matrix form.
\begin{equation}\label{recnodeloss}
    \mathcal{L}_{rec\_node} = \mathcal{L}_{node} + \sum\| \Theta_{\hat{v}}^{Con}(v) \|_1 + \sum\lambda\| \Theta_{\hat{v}}^{Con}(v) \|_2
\end{equation}

Firstly, $L1$ normalization is applied to encourage sparsity by reducing weight factors toward zero, promoting hyperedges with a minimal set of highly correlated nodes. However, $L1$ regularization can be sensitive to noise and outliers. To address this, $L2$ normalization is also introduced, providing a smoothing effect and reducing noise sensitivity. A hyperparameter \(\lambda\) is used to balance the trade-off between sparsity ($L1$) and smoothness ($L2$). This dual-regularization approach ensures robust and meaningful reconstruction of hyperedge nodes.

Meanwhile, the constraint loss is defined as:
\begin{equation}\label{constraintloss}
    \mathcal{L}_{con} =\sum_{i=1}^n\sum_{j=1}^m 
       \|f_{Con_j}(\mathcal{S}(\hat{v}_i))\|_2 
\end{equation}
where \(f_{Con_j}(\mathcal{S}(\hat{v}_i))\) evaluates constraint \(j\) on the candidate node set \(\mathcal{S}(\hat{v}_i)\) and is incorporated as a penalty term in the hyperedge node construction loss. It utilizes the upper and lower bounds of the constraints as labels and calculates the $L2$ distance as a penalty based on the constraint status of the nodes within the hyperedge. For instance, if the sum of the demands of the nodes within a hyperedge exceeds the vehicle capacity, resulting in a negative slack, it violates the capacity constraint, and the $L2$ distance between the demand sum and the capacity will be computed as a penalty. This penalty enforces compliance with constraints while allowing flexibility for soft violations.

The final encoder hypergraph loss, defined 
as \(\mathcal{L}_{hg} = \mathcal{L}_{rec\_node} + \gamma\mathcal{L}_{con}\), integrates both loss components, with \(\gamma = 0.01\) as scaling factors to balance them, ensuring neither component dominates the optimization, and enabling the model to capture structural relationships and satisfy constraints effectively within the hypergraph.

\subsubsection{MHA Node Representation Learning}
This subsection details the node representation learning module in the encoder
. The process consists of two key steps: hyperedge embedding generation and node embedding updating. Each step is described in detail below.

\paragraph{Hyperedge Embedding Generation}After selecting hyperedge nodes, the chosen nodes \(\hat{e}^{Con}(\hat{v}_i)\) are used to generate hyperedge representations under different constraints. These representations are fused with the master node features via a multi-head attention mechanism, facilitating information integration across constraints.
The hyperedge representation is determined by the projection weight coefficient \(\Theta\). To enhance this process, a gating mechanism with a predefined threshold is applied, zeroing out the weights of non-hyperedge nodes. This ensures that only relevant nodes contribute to the hyperedge features. Consequently, the embedding of the hyperedge associated with the master node \(\hat{v}_i\) is expressed as:
\begin{equation}\label{edgeembdd}
    e_{\hat{v}_i}^{Con}=\frac{H_0^{enc}(\hat{e}^{Con}(\hat{v}_i))\cdot\Bar{\Theta}_{\hat{v}_i}^{Con}(v) }{\Delta(\hat{e}^{Con}(\hat{v}_i))}
\end{equation} where
    \begin{equation}\label{thredctl}
    \Bar{\Theta}_{\hat{v}_i}^{Con}(v) = \left\{
    \begin{array}{ll}
    \Theta_{\hat{v}_i}^{Con}(v) &  v \in \hat{e}^{Con}(\hat{v}_i) \\
       1  &  v = \hat{v}_i \\ 
        0 & otherwise \\
    \end{array}
    \right.
    \end{equation}

Equation (\ref{thredctl}) implements threshold control on the weights to eliminate the influence of irrelevant nodes. The denominator of Equation (\ref{edgeembdd}) utilizes the degree of the hyperedge noted as \(\Delta\) to apply regularization.

\begin{figure*}[thb]
\centering
\includegraphics[width=1.0\textwidth]{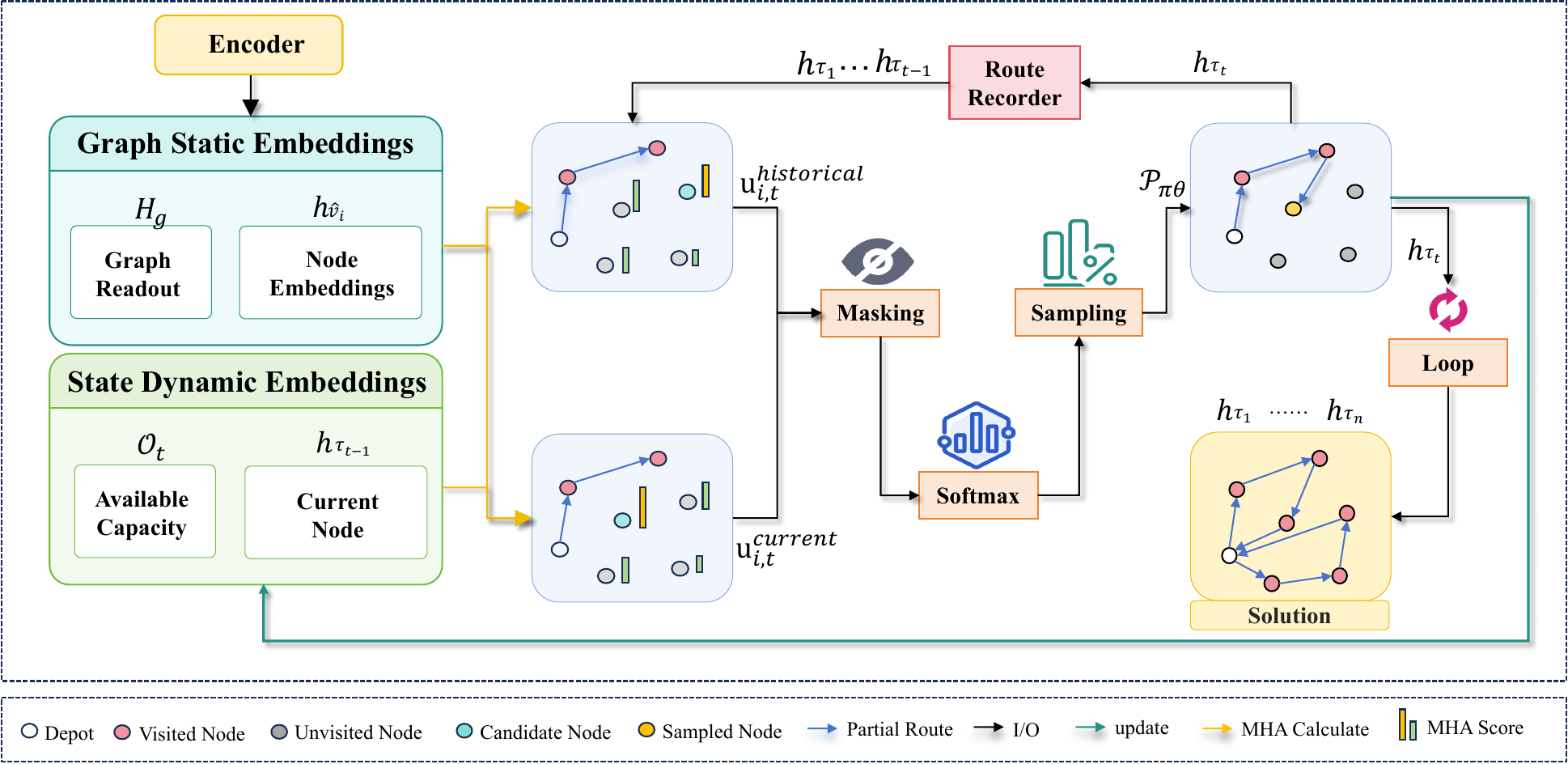}
\caption{Proposed dual-pointer attention-based decoder.}
\label{figure2}
\vspace{-5mm}
\end{figure*}
\paragraph{MHA Node Embedding Update}
As previously described, each node acts as a master node to construct hyperedges under various constraint types. Once the hyperedge representations are obtained, a multi-head attention mechanism updates the master node's embedding. This process integrates information from diverse constraint-aware hyperedges into the master node's representation, completing the graph representation learning.

To systematically explain this (see Fig.\ref{figure1}(d)), we first use linear projection matrices \(Q_{hg}^{head} \in \mathcal{R}^{\frac{d}{k} \times d}\), \(K_{hg}^{head} \in \mathcal{R}^{\frac{d}{k} \times d}\), and \(V_{hg}^{head} \in \mathcal{R}^{\frac{d}{k} \times d}, head \in \{1, \dots, k\}\) to map the master node representation \(H_0^{enc}(\hat{v}_i) \in \mathcal{R}^{d \times 1}\) and its \(j\)-th hyperedge representation \(e_{\hat{v}_i}^{Con_j} \in \mathcal{R}^{d \times 1}\) into a high-dimensional space. Here, \(k\) is the number of attention heads, and \(head\) specifies the index of each head. We then compute attention scores \(u\) by applying the dot product between the two vectors across different heads. Using these attention scores, the master node performs a weighted fusion of the hyperedge representations, as detailed in Equations (\ref{uscore}) to (\ref{nodeupdate}).
\begin{equation}\label{uscore}
    u_{\hat{v}_i}^{Con_j} = \|_{head=1}^k\frac{(Q_{hg}^{head}H_0^{enc}(\hat{v}_i))^T\cdot(K_{hg}^{head}e_{\hat{v}_i}^{Con_j})}{\sqrt{d/k}}
\end{equation}
The equation (\ref{uscore}) computes the attention score between the master node \( \hat{v_i} \) and one of its hyperedges \( \hat{e}^{Con_j}(\hat{v}_i) \). The symbol \( || \) denotes the concatenation operation, which merges the attention scores from various heads. The attentive score \( u_{\hat{v}_i}^{Con_j} \in \mathcal{R}^k \) is a weight vector that illustrates the importance of a specific constrained hyperedge, referred to as superscript \( ^{Con_j} \), in updating the embedding of the master node.
\begin{equation}\label{nodeupdate}
    h_{\hat{v}_i} =\|_{head=1}^k\sum_{j=1}^mSoftmax\left( u_{\hat{v}_i}^{Con_j}\right)\cdot V_{hg}^{head}e_{\hat{v}_i}^{Con_j}
\end{equation}

Equation (\ref{nodeupdate}) uses the \textit{Softmax} function to normalize hyperedge attention scores \( u \in \mathcal{R}^k \) across different constraints. The normalized weights are used for weighted fusion of hyperedges, mapped via the projection matrix \(V_{hg}^{head} \in \mathcal{R}^{\frac{d}{k} \times d}\) over constrained hyperedges 1 to \(m\). Outputs from all attention heads are then aggregated to form the fused master node representation \( h_{\hat{v}_i} \in \mathcal{R}^{d \times 1} \) (Fig.\ref{figure1}(d)). Finally, average pooling is applied to all fused master nodes to obtain the hypergraph's overall representation for the decoder, given by \(H_g = \frac{\sum_{i=1}^n h_{\hat{v}_i}}{n}\).

\subsection{Dual-Pointer Decoder}
Decoders in previous studies~\cite{belloNeuralCombinatorialOptimization2017,nazariReinforcementLearningSolving2018,koolAttentionLearnSolve2019,kwonPOMOPolicyOptimization2020,leiSolveRoutingProblems2022} relied solely on current state features, using node representations from the encoder for stochastic sampling. Violated or visited nodes were masked, and the depot was selected if no valid nodes remained. While solutions were generated iteratively, this approach suffered from error propagation, as suboptimal decisions at one step impacted subsequent steps, complicating learning. To overcome this, we propose a dual-pointer decoder that combines current and historical state features. A route recorder tracks partial solutions, enabling attention over global nodes with fused probability distributions from both pointers. This design improves fault tolerance, reducing error propagation and enhancing robustness and performance.

\subsubsection{Context Embedding}
To maintain the continuity of historical states, we propose a route recorder module that records the representation data of previously visited nodes and incorporates it through a residual connection to form the embedding of the partial solution. This mechanism effectively emphasizes the representation of completed nodes during the attention calculation process. Compared to approaches that rely solely on the previous node, this design is more conducive to guiding decisions toward peripherally related nodes. The primary advantage lies in its ability to introduce perturbations into the decision-making process, thereby reducing the likelihood of converging to local optima. Consequently, this approach enables a more comprehensive consideration of context nodes. Accordingly, the embedding of context nodes is divided into two components, as defined in Equation (\ref{curentembed}) and Equation (\ref{hisembed}).
\begin{equation}\label{curentembed}
    h_{c}^{current} = \left\{ \begin{array}{ll}
         FF\left(H_g  \| h_{v_0}  \| \mathcal{O}_t \right) & t = 0\\
         FF\left( H_g  \| h_{\tau_{t-1}} \| \mathcal{O}_t \right) & t > 0\\
         
    \end{array}\right.
\end{equation}

\begin{equation}\label{hisembed}
    h_{c}^{historical} = \left\{ \begin{array}{ll}
         FF\left(H_g  \| h_{v_0}  \| \mathcal{O}_t \right) & t = 0\\
         FF\left( H_g  \| h_{\tau_{t-1}} + ... + h_{\tau_{t1}} \| \mathcal{O}_t \right) & t > 0\\
         
    \end{array}\right.
\end{equation}
\(FF\) denotes the feed forward layer, while \(\mathcal{O}_t\) represents the state of the vehicle at timestep \(t\), specifically its remaining capacity. At the initial timestep (\(t = 0\)), the two context vectors are identical. However, as the solution is iteratively constructed, \(h_c^{historical}\) progressively incorporates the representations of visited nodes into the context embedding, enabling it to capture historical information dynamically.

\subsubsection{Dual Probability Calculation}
Following the approach in Equation (\ref{uscore}), we utilize a multi-head attention mechanism to calculate attention scores between the two context embeddings and all node embeddings. These scores are then used for stochastic sampling based on a probability distribution. Rather than sampling independently from the two pointers, we compute a weighted average of their probability distributions. This method smooths the sampling process and increases the chances of selecting nodes that a single pointer might overlook. As a result, it reduces the risk of the decoder converging to a local optimum, as shown in Fig.~\ref{figure2}. The detailed computation is presented in Equations (\ref{head-coe-cus}) and (\ref{ptheta}).
    \begin{equation}\label{head-coe-cus}
    u_{i,t}^{c_i} = \begin{array}{ll}
\frac{\left(\boldsymbol{Q}_{c_i}h_c^{c_i}\right)^T
        \left(\boldsymbol{K}_{c_i}h_{\hat{v}_i}^t \right)}{\sqrt{d_h}}   & \hat{v}_i \neq \tau_{t^\prime} , \forall t^\prime < t\\
         \\
    \end{array} 
\end{equation}
Here \(c_i = \{current,historical\}\) indicates the index of pointers and context embeddings. For each policy, we use separate weight matrices for query and key projections in attention calculation.  \(h_{\hat{v}_i}^t\) represents the candidate action node that can be selected (after masking unavailable nodes) at timestep \(t\), while \(d_h\) is the feature dimension on each attention head for controlling variance and preventing gradient vanishing.
Equation (\ref{head-coe-cus}) derives the attention compatibility \(u\) for current and historical states individually, considering the available nodes at step \(t\). Equation (\ref{ptheta}) calculates the probability distribution for sampling. To enhance the distinction between nodes, activation and scaling are applied to the two attention vectors separately, constraining their values within \([-Clip, Clip]\) (with \(Clip = 10\)), following prior works \cite{koolAttentionLearnSolve2019,leiSolveRoutingProblems2022,wang2024gase}. The final probability \(p_{\pi_\theta}\) combines the attention results from both pointers, guiding the decoder to randomly select a node as the action node at step \(t\).
\begin{equation}\label{ptheta}
    p_{\pi_\theta}(\tau_t|s, \tau_{t^\prime}, \forall t^\prime < t) = Softmax(\sum_{c_i} Clip\cdot tanh (u_{i,t}^{c_i}))
\end{equation}
The decoding operation must be executed in a loop for a certain number of steps, referred to as \(t\) until all nodes have been visited. Notably, the value of \(t\) is usually much larger than the number of nodes, $n$. The reason is that in routing problems, vehicles often select the depot as intermediate nodes to complete sub-paths owing to the constraints, which increases the number of steps required. Moreover, the attention score for infeasible nodes is set as \(-\inf\), ensuring the probability of being selected is \(0\) during the masking operation.

\subsection{REINFORCE Self-Critic Training}
We adopt reinforcement learning to train the proposed end-to-end model. The primary objective is to expose the model to a large number of problem instances drawn from the same distribution, allowing it to learn generalizable patterns and enhance its decision-making performance. Through training, the model learns to increase the likelihood of making optimal decisions. When encountering similar problems characterized by specific features, the model can select a favorable combination of decision actions, as these actions have been optimized during the training process to achieve superior solutions. 

To accurately estimate the expected reward for sampling a batch of data, we employ Monte Carlo estimation. The estimated reward is then used as the loss function for reinforcement learning, formulated as \(\mathcal{L}_{rl}({\pi_\theta}|\mathcal{G}) = \mathbb{E}_{\tau\sim p_{\pi_\theta}(\tau | \mathcal{G})}r(\tau)\). During training, we utilize the REINFORCE algorithm \cite{zhang2021sample} and evaluate the model against a self-critical baseline to further stabilize and improve the learning process.
\subsubsection{REINFORCE with Greedy Baseline Rollout}
During the training process, strategies that outperform the baseline are identified as effective strategies. To leverage this, we adopt a variant of the actor-critic approach called the self-critic mode. In this approach, the same model is used for both the actor and the critic, but with different decoding strategies to evaluate the quality of the actor model. Specifically, we use the parameters of the best-performing actor model to date as the parameters for the critic (baseline) model. For the baseline evaluation, a greedy decoding strategy is employed, which consistently selects the action with the highest probability. This simplifies the evaluation process and provides a stable and reliable reference point.

To further streamline computation, all probabilities are converted into logarithmic probabilities, effectively reducing computational complexity. With these considerations, the policy gradient \cite{suttonPolicyGradientMethods1999} can be formulated as:
\begin{equation}
    \nabla \mathcal{L}_{rl} \left( \pi_\theta | \mathcal{G} \right) = \mathbb{E}_{\tau\sim p_{\pi_\theta}(\tau | \mathcal{G})}[(r(\tau) - b_{\pi_{\theta^\prime}}(\mathcal{G})) \nabla_\theta \log p_{\pi_{\theta}}(\tau | \mathcal{G})]
\end{equation}
In this context, \(\pi_{\theta^\prime}\) refers to the parameters of the previous optimal model. As training progresses, if the reward from the actor network surpasses that of the critic network, the actor's policy is reinforced. When the actor's performance on the validation set significantly exceeds that of the critic network, the parameters of the critic network are updated to match those of the current actor network. This continuous updating process allows the critic network to help the actor consistently perform better than greedy rollout, facilitating gradual convergence during training.

\subsubsection{Self-Critic Training}
The hypergraph encoder incorporates two types of loss: the hyperedge node reconstruction loss and the hyperedge constraint loss, as described in Section 4.2. When applying self-critic methods in reinforcement learning to update the policy gradient of network parameters during training, the encoder's parameters are updated separately using a combined loss function. It is worth noting that, when employing the Adam optimizer for network parameter updates, the decoder's parameters are updated once per batch of sampled data. 

However, during our experiments, we observed that frequent updates to the hypergraph encoder's parameters could lead to non-convergence due to the large volume of data involved in reinforcement learning training. To address this issue, we opted to update the hypergraph network's parameters only once after completing the training for an entire epoch. This ensures that all instance features are fully learned before the parameters are adjusted. The detailed training procedure is outlined in Algorithm \ref{alg1}.
\begin{algorithm}[htb]
	\renewcommand{\algorithmicrequire}{\textbf{Input:}}
	\renewcommand{\algorithmicensure}{\textbf{Output:}}
    \caption{Asynchronous REINFORCE Algorithm}\label{alg1}
\begin{algorithmic}[1]
    \REQUIRE Actor policy network $\pi_\theta$; Baseline policy network $\pi_\theta^b$; Training epochs $E$; Batch size $B$; Steps per epoch $T$; Significance level $\epsilon$.
    \STATE Initialization: $\theta, \theta^b \gets$ Xavier init $\theta$;
    \FOR{ $e$ in $1,\ldots,E$}
    \FOR{ $t$ in $1,\ldots,T$}
    \STATE $\mathcal{G}_i \gets RandomInstance(), \forall i \in \{1,\ldots,B\}$
    \STATE Compute $\mathcal{L}_{hg}$
    \STATE $\tau_i \gets SampleSolution(\mathcal{G}_i, p_\theta), \forall i \in \{1,\ldots,B\}$
    \STATE $\tau_i^b \gets GreedyRollout(\mathcal{G}_i, p_\theta^b), \forall i \in \{1,\ldots,B\}$
    \STATE Compute $r(\tau_i | \mathcal{G}_i), r(\tau_i^b | \mathcal{G}_i) $ through Eq. (\ref{rewardcal})
    \STATE $\nabla\mathcal{L}_{rl} \gets \sum_{i=1}^B (r(\tau_i) - r(\tau_i^b))\nabla_\theta \log p_{\pi_\theta}(\tau_i | \mathcal{G}_i)$
    \IF{$\theta_{rl} \in \theta$}
    \STATE $\theta_{rl} \gets Adam(\theta_{rl}, \nabla_\theta\mathcal{L}_{rl})$
    \ENDIF
    \ENDFOR
    \IF{$\theta_{hg} \in \theta$}
    \STATE $\theta_{hg} \gets Adam(\theta_{hg}, \nabla_\theta\mathcal{L}_{hg})$
    \ENDIF
    \IF{OneSidedPairedTTest($p_\theta, p_\theta^b$) $< \epsilon$}
    \STATE $\theta^b \gets \theta$
    \ENDIF
    \ENDFOR
    \ENSURE Parameter set $\theta$ of the actor network       
\end{algorithmic}  
\end{algorithm}
This asynchronous parameter update mechanism significantly improves overall model performance. First, the stable, high-quality graph feature representations learned by the encoding module at the epoch level provide more reliable inputs to the decoding module. Due to the complexity of the graph structure information processed by the hypergraph network, frequent parameter updates can lead to abrupt changes before the model fully captures the graph structure features and constraint relationships. Such instability may result in over-smoothing of graph node representations, reducing the discriminability of attention mechanisms during decoding and hindering the convergence of reinforcement learning. From the perspective of reinforcement learning, this design separates two sources of feedback with different statistical properties. The decoder receives batch-level route rewards and updates the policy so that trajectories outperforming the greedy self-critical baseline increase their log-probability, while inferior trajectories are penalized. The hypergraph encoder, in contrast, is guided by reconstruction and constraint losses accumulated over an epoch. This separation reduces the interference between sparse route-level reward signals and representation-level hyperedge reconstruction signals.

Meanwhile, the decoder processes samples in batches, providing immediate feedback that enables timely parameter adjustments. This increases the likelihood of adopting more effective reward strategies, accelerating the model's search for local optimal solutions. In the early stages of training, this approach helps the decoding module quickly learn basic output strategies while simultaneously offering positive feedback to guide parameter updates in the encoding module. As training progresses, the hyperedges derived from the learned representations of the encoding module become better aligned with the problem requirements, effectively narrowing the search space during decoding and facilitating faster convergence of the overall training process.

\subsection{Complexity Analysis}

We analyze the main computational cost of the proposed encoder-decoder framework. Let \(N=n+1\) be the number of nodes including the depot, \(d\) the hidden dimension, \(L_g\) the number of pairwise graph-attention layers before hypergraph construction, \(m\) the number of constraint types, \(\bar{\Delta}\) the average number of retained nodes in a hyperedge after thresholding, and \(T\) the number of decoding steps.

The low-order graph encoding stage first projects node features and then applies pairwise graph attention. Its dominant cost is
$
C_{\mathrm{pair}}=O\!\left(L_g(N^2d+Nd^2)\right)$,
where \(O(N^2d)\) comes from pairwise attention score computation and value aggregation, and \(O(Nd^2)\) comes from feature projections.

The hypergraph reconstruction stage constructs constraint-oriented hyperedges for each master node. For \(m\) constraint types, scoring all master-candidate node pairs requires
$
C_{\mathrm{score}}=O(mN^2d)$.
After thresholding, each hyperedge contains \(\bar{\Delta}\) nodes on average, so hyperedge aggregation and constraint-aware representation fusion require
$
C_{\mathrm{hg}}=O(mN\bar{\Delta}d)$.
This term also shows the advantage over clique expansion: converting a degree-\(\Delta\) hyperedge into pairwise edges would introduce \(O(\Delta^2)\) pairwise relations and remove the explicit identity of the constraint-induced group, whereas direct hyperedge aggregation preserves the group relation with cost linear in \(\bar{\Delta}\).

The dual-pointer decoder evaluates candidate-node scores at each construction step by using the current node state and the route-history context. Under greedy decoding, its dominant cost is
$
C_{\mathrm{dec}}=O(TNd)$,
or \(O(TNd+T^2d)\) if the route-history context is fully recomputed rather than incrementally updated. Sampling, beam search, or local search would multiply or add to this decoding cost.

Therefore, the dominant per-instance cost of the proposed model is:
\begin{equation}
    C_{\mathrm{total}}
=
O\!\left(
L_g(N^2d+Nd^2)
+
mN^2d
+
mN\bar{\Delta}d
+
TNd
\right)
\end{equation}

For mini-batch training with batch size \(B\) and \(A\) augmented rollouts, the forward cost scales as \(O(BA\,C_{\mathrm{total}})\). The main memory cost comes from dense pairwise attention and hyperedge reconstruction coefficients, i.e., \(O(N^2+mN^2)\), plus the retained sparse hyperedge incidence \(O(mN\bar{\Delta})\).

\section{Experiment}
We conducted a series of experiments to evaluate the effectiveness of our proposed constraint-oriented hypergraph model. Using three datasets, we analyzed the model from three key perspectives to address the following research questions:

RQ1. Solution Quality: Does our model outperform state-of-the-art methods in solving the VRP and its variants? Specifically, we evaluated improvements in solution quality, computational efficiency, and the model's capability to handle online problem-solving scenarios.

RQ2. Ablation Study: How do individual components of the model contribute to its overall performance? We investigated the effects of the data augmentation module, hypergraph module, and double-pointer decoding module.

RQ3. Hyperparameter Sensitivity: How do hypergraph reconstruction parameters and hyperedge mapping regularization coefficients influence the model's performance? We conducted sensitivity analyses to explore the impact of these hyperparameters.

These research questions guided a comprehensive evaluation of our model's performance and characteristics. In the following sections, we detail the experimental setup and provide an in-depth analysis of the results.
\subsection{Experiment Setup}
\paragraph{Datasets and Parameters}
Building on prior research, we utilized both randomly generated datasets and the CVRPLIB benchmark to evaluate the effectiveness of our model in solving the CVRP and its variants. For the randomly generated datasets, we trained three models corresponding to problem sizes of 20, 50, and 100 nodes. Node demands were randomly sampled within the range of 1 to 9, while vehicle capacity constraints were set to 30, 40, and 50, respectively, based on the problem size.

For the CVRPLIB dataset, we employed pre-trained models to address problems of varying scales. Specifically, the model trained on 50-node problems was used to solve instances with fewer than 60 nodes, while the model trained on 100-node problems was applied to larger instances. This approach ensured the scalability and adaptability of the model across different problem sizes.

Furthermore, we implemented and evaluated diverse graph-based end-to-end models on the CVRPTW under soft time window constraints. Each node's service time constraint was defined with a uniformly distributed random start time \(T_{start}\) combined with a randomly generated service time \(T_{service}\) and accepted service duration \(T_{window}\). Consequently, the time window of customers is represented as \([T_{start}, T_{start}+T_{window}]\) while vehicles stayed at each node during service time \(T_{service}\). Vehicle speed was maintained constant at $1$ unit, and deviations from the designated time windows incurred penalties proportional to temporal discrepancies: a penalty coefficient of $0.5$ was applied for early arrivals (relative to the start time) and $2.0$ for late arrivals. The optimization objective aimed to minimize the total sum of vehicle travel time and incurred penalties across all routes. This formulation ensures a balanced trade-off between operational efficiency and temporal constraint adherence while preserving computational tractability. 

For parameter selection during training, we used $640,000$ instances to train the models with $20$ and $50$ nodes, while the model with $100$ nodes was trained on $128,000$ instances due to time and memory constraints. Both the validation and test sets consisted of $1,000$ instances each, and the results were averaged over these $1,000$ instances. The learning rate was set to \(10^{-4}\), with a decay factor of $0.96$ for Adam Optimizer~\cite{kingma2014adam}. The batch size was $128$ for the $20-$ and $50-$node models and $64$ for the $100-$node model. The hidden feature dimension of the neural network was set to $256$, with $8$ attention heads, a dropout rate of $0.1$, and $200$ training epochs to ensure stability and convergence. 

All experiments were conducted on an Nvidia RTX A6000 GPU with 48GB of memory, paired with an Intel Xeon Silver 4126 CPU running at 2.10GHz.

\paragraph{Baseline Models}
We categorized the baseline models into three groups for comparison with our proposed model. The three categories are: general solvers, sequential end-to-end models, and graph-based end-to-end models. The general solvers include Gurobi~\cite{Gurobi2024}, LKH3~\cite{helsgaun2017extension}, and Google OR-Tools~\cite{ortools}. For end-to-end models, we selected classic SOTA models, graph-based models, and recent neural routing solvers with heavy-decoder, ensemble local-policy, and light-decoder designs.

Below, we provide an introduction to the learning-based baselines:
\begin{itemize}
\item  Pointer Networks (PtrNets)~\cite{nazariReinforcementLearningSolving2018}: A classic seq2seq model, which integrates LSTM and dot-product attention, is capable of efficiently deriving solutions for the VRP within a short time.

\item Attention Model (AM)~\cite{koolAttentionLearnSolve2019}: The transformer-based classic SOTA model, utilizing a seq2seq framework, has been extensively applied to solve the VRP and its variants. It has established itself as a foundational milestone, significantly influencing subsequent research in this domain.

\item  POMO~\cite{kwonPOMOPolicyOptimization2020}: The enhanced algorithm of Attention Model (AM) integrates data augmentation techniques and a multi-start parallel strategy. It has been widely applied to solving problems such as the Traveling Salesman Problem and CVRP.

\item LEHD~\cite{luo2023lehd}: A heavy-decoder neural solver that reduces reliance on a static encoder and shifts relation modeling into the dynamic decoding stage, thereby improving cross-scale generalization for routing problems.

\item ELG~\cite{gao2023towardslocalglobal}: An ensemble framework with a transferable local policy that combines local and global information to improve the generalization of neural VRP solvers.

\item ReLD~\cite{huang2025rethinking}: A recent light-decoder-based solver that reconsiders decoder context design and augmentation for efficient neural routing.

\item  E-GAT~\cite{leiSolveRoutingProblems2022}: The model based on Graph Attention Networks, incorporating edge features, represents the SOTA among graph-based approaches for solving VRP. Its encoder employs GAT with residual connections, while the decoder follows a design similar to that of the AM.

\item  GASE~\cite{wang2024gase}: This work builds upon AM and GAT by integrating additional sampling policy into the graph node embeddings to better regulate the attention mechanism, thereby improving solution quality.
\end{itemize}

\subsection{Experiment Results and Analysis}

\begin{table*}[thb!]
\caption{Comparison with baseline methods on random CVRP instances. The symbol ``-'' in the Gurobi row indicates that no result was available within the time limit; Gurobi and LKH3 outcomes are reproduced from previous studies~\cite{koolAttentionLearnSolve2019,kwonPOMOPolicyOptimization2020,leiSolveRoutingProblems2022}. The symbol ``*'' in the Time column denotes the total runtime of a single-threaded sequential OR-Tools run with a 1-second search limit per instance over 1,000 test cases. Boldface denotes SOTA performance among learning-based methods.} 
        \label{basicresult}
	\centering
       \resizebox{\linewidth}{!}{
	\begin{tabular}{lccccccccc}
		\toprule
		\multirow{2}*{Model}& \multicolumn{3}{c}{CVRP20} & \multicolumn{3}{c}{CVRP50}& \multicolumn{3}{c}{CVRP100}\\ 
		\multirow{2}*{}& Objective & Gap(\%) & Time & Objective & Gap(\%) & Time & Objective & Gap(\%) & Time\\
	
        \midrule 
		Gurobi\cite{Gurobi2024}       & 6.10 & 0.00 & -  & -     & -    & -  &   -   &  -   &  -  \\
		LKH3\cite{helsgaun2017extension}          & 6.14 & 0.58 & 2h & 10.38 & 0.00 & 7h & 15.65 & 0.00 & 13h  \\
		OR-Tools\cite{ortools}     & 6.18 & 1.31 & *1000s  & 10.98 & 5.81 & *1016s  & 17.09 & 9.20 &  *1054s  \\
        \midrule
        PtrNet\cite{nazariReinforcementLearningSolving2018}       & 6.59 & 8.03 & \textbf{0.11s}  & 11.39 & 9.78 & \textbf{0.16s}  & 17.23 & 10.12 & \textbf{0.32s} \\
           
		AM model\cite{koolAttentionLearnSolve2019}     & 6.40 & 4.97 & 1s  & 10.98 & 5.86 & 3s & 16.80 & 7.34  & 8s  \\
        
        AM $Sampling$ \cite{koolAttentionLearnSolve2019}     & 6.24 & 2.30 & 3m  & 10.59 & 2.02 & 7m & 16.16 & 3.26  & 30m  \\
            
            POMO \cite{kwonPOMOPolicyOptimization2020}         &6.35& 4.09 &  1s&  10.74&  3.52 &    1s &  16.15 & 3.19  &3s  \\
            
            POMO$\times 8 Aug$\cite{kwonPOMOPolicyOptimization2020}& 6.14 & 0.65 &  5s&  10.42&  0.45 &    26s &  \textbf{15.73} & \textbf{0.32}  &2m  \\
            LEHD Greedy~\cite{luo2023lehd} & 6.36 & 4.26 & 2s & 10.92 & 5.41 & 2s & 16.22 & 4.59 & 30s \\
            ELG$\times 8$Aug ensemble 0.4 Local Size~\cite{gao2023towardslocalglobal} & 6.35 & 4.16 & 4s & 10.47 & 1.06 & 10s & 15.84 & 2.13 & 3.18m \\
            ReLD$\times 8$Aug~\cite{huang2025rethinking} & 6.15 & 0.89 & 3s & 10.43 & 0.66 & 9s & 15.80 & 1.87 & 1.34m \\
            \midrule
		E-GAT\cite{leiSolveRoutingProblems2022}        & 6.26 & 2.60 & 2s & 10.80  & 4.05 & 7s & 16.69 & 6.68 & 17s \\
        GASE\cite{wang2024gase}         & 6.21 & 1.80 & 1.7s &10.74  & 3.46 & 3s & 16.37  & 4.60   &  6.6s \\
        
        \midrule
        
		{\bf Our approach} &  6.14 & 0.65  & 3s  &  10.65 & 2.60  &  8s  & 16.35 & 4.47 & 18s \\

        {\bf Our approach $\times 8 Aug$} &  \textbf{6.13} & \textbf{0.49}  & 6s  &  \textbf{10.40} & \textbf{0.19}  &  28s  & 15.77 & 0.76 & 2m \\
		\bottomrule
	\end{tabular}
 
 }
\end{table*}
This subsection analyzes the experimental results by addressing the research questions posed earlier through three parts of the experiment: the performance of the model on benchmark datasets, the effectiveness of individual model components, and the impact of hyperparameter variations on model performance.
\subsubsection{Model Performance}
We first evaluated the trained model on datasets where node coordinates were uniformly and randomly generated within the range [0, 1], and compared its performance with other SOTA methods on CVRP. As shown in Table \ref{basicresult}, the ``Objective'' column records the average route length over 1,000 test cases. The results for Gurobi, LKH3, and other learning-based baselines are taken from referenced work~\cite{leiSolveRoutingProblems2022}. The hypergraph-based model demonstrates outstanding performance on small- and medium-scale datasets, establishing new SOTA results for small- and medium-scale problems among all end-to-end models. We attribute this to the fact that, on small- and medium-scale datasets, hyperedges constrain the decision space to a subset of nodes satisfying hyperedge constraints. Compared to other SOTA methods, which search across the entire instance's node space, the hypergraph model operates in a significantly smaller sampling space, making it more likely to achieve better results within a limited number of stochastic samples. The proposed model achieves performance improvements ranging from 0.82\% to 7.38\% compared to existing end-to-end models. Furthermore, its exceptionally short inference time enables effective online decision-making for problems with similar distributions, without compromising efficiency. 

To address recent SOTA comparisons more explicitly, Table~\ref{basicresult} additionally includes LEHD Greedy, ELG with 8-fold augmentation and ensemble local policy, and ReLD with 8-fold augmentation. These methods represent complementary recent directions: heavy-decoder relation reconstruction, local-global ensemble policies, and light-decoder redesign. Under this expanded comparison, our approach with 8-fold augmentation achieves gaps of 0.49\% and 0.19\% on CVRP20 and CVRP50, respectively, and remains competitive with recent neural solvers while using a constraint-oriented hypergraph encoder. On CVRP100, however, POMO with 8-fold augmentation remains stronger in gap, whereas our model remains competitive but incurs a larger inference cost. This pattern indicates that the proposed hypergraph encoder is most effective when constraint-compatible customer groups can be reliably reconstructed from the training scale, while larger instances require stronger decoding or cross-scale training to fully exploit the representation.
\begin{figure*}[!thb]
\centering
\includegraphics[width=0.99\textwidth]{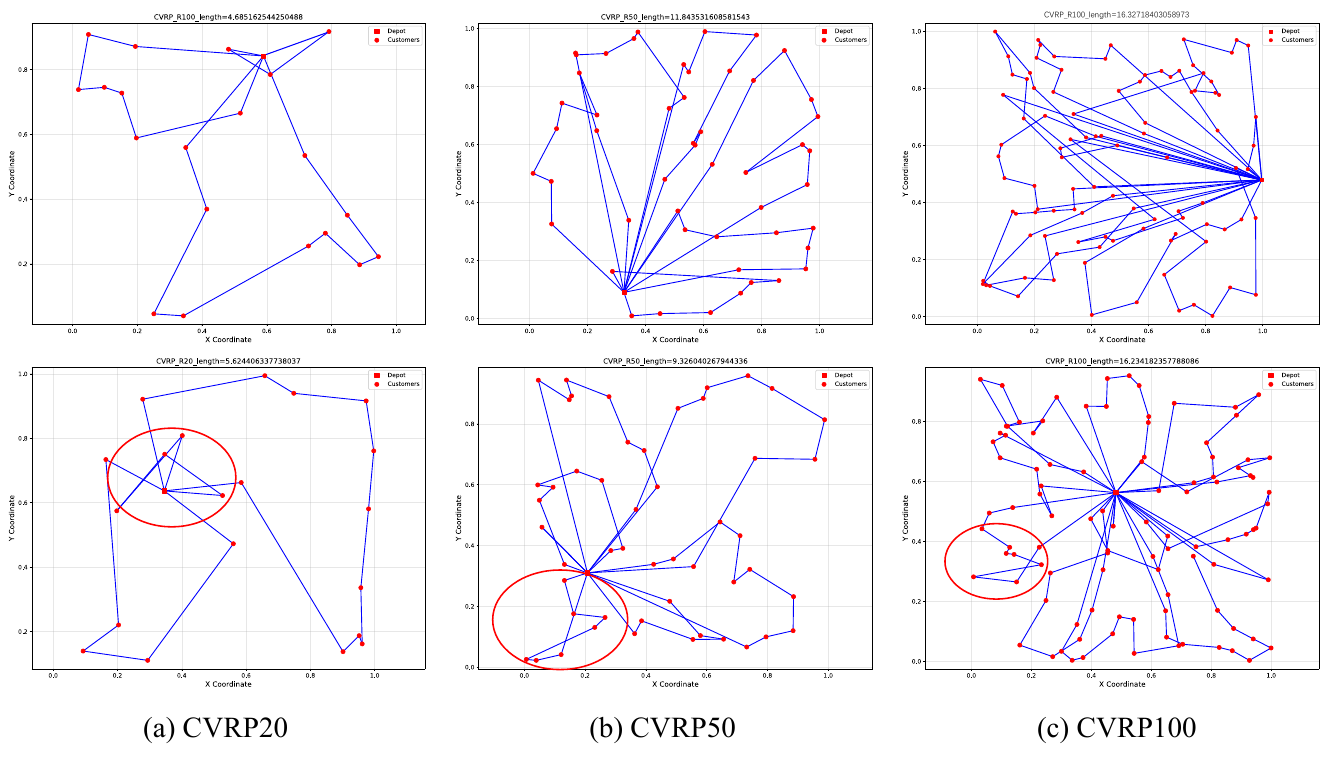}
\caption{Visualization on randomly generated CVRP datasets. The first row demonstrates that the hyperedge constraint learning effectively captures optimized vehicle capacity patterns, with the sub-routes exhibiting no discernible suboptimal path segments internally. In contrast, while the second row shows successful learning of vehicle capacity patterns through hyperedge constraints, the red-circled portions within the sub-routes reveal apparent non-optimal routing configurations that require further optimization.}
\label{figure_visual}
\end{figure*}

We also visualized the generated solutions to identify the performance bottlenecks of the proposed hypergraph model. As illustrated in Fig.~\ref{figure_visual}, the hyperedge constraints within the hypergraph framework effectively cluster nodes into subpaths for a single vehicle, forming a cyclical pattern. The decoder then resolves these subpaths, effectively decomposing the VRP into multiple TSPs that satisfy capacity constraints, thereby reducing the complexity of the solution process. During decoding, the first row of Fig.~\ref{figure_visual} demonstrates ideal scenarios where the capacity constraints are evenly distributed, and no significant suboptimal decision subsequences are present. In contrast, the second row exhibits clear suboptimal decision sequences, as indicated by the crossing of subpaths within the red circles, caused by the greedy strategy employed during inference. Common approaches to address this performance bottleneck include expanding the search space for each decision step (e.g., beam search), utilizing stochastic sampling to perform multiple inferences and select the optimal result (Sampling), or applying heuristic methods such as local search to enhance the obtained solutions. 
Although these approaches enhance the quality of the solutions, they inevitably increase inference time. For online problems, this often translates to longer solving times, requiring a trade-off between inference time and solution quality based on the specific application scenarios.
\begin{table}[tbh]
\caption{Comparison results on random CVRPTW }
        \label{VRPTWresult}
	\centering       
	\begin{tabular}{lcccccc}
		\toprule
		\multirow{2}*{Model} & \multicolumn{2}{c}{CVRPTW20} &  \multicolumn{2}{c}{CVRPTW50}& \multicolumn{2}{c}{CVRPTW100}\\ 
		& Objective  & Time & Objective & Time & Objective  & Time\\
			
        \midrule 
           
		E-GAT        & 14.39 &  2s & 36.8  &  7.8s & 76.87 &  18.1s \\
        GASE         & 13.62 &   1.8s & 33.79 &   3.2s &  69.63 &     8.3s  \\
        		
        \hline 
            \hline 
		{\bf Our approach} &  9.27  & 3.2s  &  16.27  &  9.7s  & 44.93  & 23s \\
		\bottomrule
	\end{tabular}
\end{table}

Furthermore, we applied the proposed hypergraph model and other comparative graph-based models to the randomly generated VRPTW. Notably, the hypergraph model outperformed alternative graph architectures, demonstrating its superior capability in capturing high-order node interactions for sequential decision-making under time window penalties. Given the stochastic generation of time windows, where single-step decision errors may propagate into substantial delays, our hyperedge construction explicitly incorporates next-node arrival status as a loss term to mitigate temporal constraint violations. As evidenced by the experimental results in Table \ref{VRPTWresult}, this targeted hyperedge design for time window nodes effectively reduces penalty accumulation while maintaining computational efficiency. The improvement on CVRPTW is substantially larger than that on standard CVRP: for CVRPTW100, the objective is reduced from 76.87 with E-GAT and 69.63 with GASE to 44.93 with the proposed model. This supports the motivation that time-window feasibility is not a purely pairwise spatial relation. A customer that appears locally attractive by distance can still become unfavorable after considering the accumulated service time and the temporal status of other customers in the same partial route. The constraint-oriented hyperedge therefore supplies a useful group-level signal to the decoder. 

\begin{table*}[tb]
\caption{Model performance on CVRPLIB datasets across different problem scales. The boldface entries denote the best results achieved by comparative models. The symbol "-" indicates entries that are not applicable or not reported. The Gap column quantifies the performance discrepancy between the current model and the optimal solution.}\label{vrplibresult}
	\centering
        \resizebox{\linewidth}{!}{
	\begin{tabular}{lcccccccc}
		\toprule
		\multirow{2}*{Data instance} & \multirow{2}*{Number of nodes}& \multicolumn{1}{c}{Optimal Solution} &  \multicolumn{2}{c}{Our Approach}& \multicolumn{2}{c}{E-GAT}&\multicolumn{2}{c}{POMO$\times 8Aug$}\\ 
		
		& & Objective & Objective & Gap(\%) & Objective & Gap(\%) &  Objective & Gap(\%)    \\
		\midrule
 
            A-n44-k6  & 43  & 937   & 985  & 5.12   & \textbf{984}     & \textbf{5.01}  & 997   & 6.40 \\
		A-n45-k6  & 44  & 944   & \textbf{952}  & \textbf{0.85}   & 984     & 4.23  & 993  & 5.52 \\
           
		A-n63-k10 & 62  & 1314  & 1375  & 4.64    & 1519    & 15.60 & \textbf{1329}  & \textbf{1.16} \\
            A-n64-k9  & 63  & 1401  & 1549 & 10.56   & 1659    & 18.40 & \textbf{1430}  & \textbf{2.10} \\
            A-n69-k9  & 68  & 1159  & 1203  &  3.80   & 1264    & 9.05  & \textbf{1199}  &\textbf{3.47} \\
            B-n34-k5  & 33  & 788   & \textbf{793}   & \textbf{0.63}    & 812     & 3.04  & 903   & 14.66 \\
            B-n35-k5  & 34  & 955   & \textbf{983}   &  \textbf{2.93 }  & 986     & 3.24  & 1005  & 5.28 \\
            B-n45-k6  & 43  & 678   & \textbf{707}   &  \textbf{4.28}  & 729     & 7.52  & 737   & 8.73\\
            B-n51-k7  & 50  & 1032  & 1121  & 8.62   & \textbf{1045}    & \textbf{1.25}  & 1088  & 5.48\\
            P-n50-k8  & 49  & 631   & 665   & 5.39    & \textbf{655}     & \textbf{3.80}  & 661   & 4.83 \\
            P-n51-k10 & 50  & 741   & 761   & 2.70   & 811     & 9.44  & \textbf{759}  & \textbf{2.43} \\
            P-n70-k10 & 69  & 827   & \textbf{862}   &  \textbf{ 4.23}  & 865     & 4.59 & 867   & 4.79 \\
            E-n101-k14  & 100  & 1067   &  1126  &  5.53  & 1163     & 8.96  & \textbf{1116}   & \textbf{4.22} \\
  M-n101-k10  & 100  & 820   &  \textbf{827}  & \textbf{0.85}   &  873     & 6.46  & 842   & 2.98 \\
  M-n121-k7  & 120  & 1034  & 1141 & 10.35  & 1144    & 10.63 & \textbf{1110}   & \textbf{7.35} \\

  M-n151-k12 &	150	& 1015	& 1120 & 10.34 &	1106&	8.96&	\textbf{1070}&	\textbf{5.44}\\
            \hline
            \hline
  {Average Gap}  & - & - & - & \textbf{5.01} & - & 7.51  & - & 5.30  \\
		\bottomrule
	\end{tabular}
 }
\end{table*}

\begin{table*}[tb]
\caption{Analysis on XML100 customer-distribution subsets. All methods use 100-node pretrained models and are evaluated on XML100 instances with the same node scale but different geometric distributions. ``Mixed'' denotes the random-clustered customer distribution in XML100.}
\label{tab:xml100_scalability}
\centering
\resizebox{0.92\linewidth}{!}{

\begin{tabular}{lccccc}
\toprule
Customer distribution & Instances & \multicolumn{2}{c}{POMO} & \multicolumn{2}{c}{Our approach} \\
 & & Gap(\%) & Gap Std & Gap(\%) & Gap Std \\
\hline
Clustered & 3322 & 10.59 & 9.11 & \textbf{10.11} & \textbf{8.32} \\
Mixed & 3318 & 8.75 & 7.83 & \textbf{8.41} & \textbf{6.77} \\
Clustered, 8Aug & 3322 & 6.62 & 5.05 & \textbf{6.48} & \textbf{4.83} \\
Mixed, 8Aug & 3318 & 5.77 & 4.06 & \textbf{5.55} & \textbf{3.73} \\
\bottomrule
\end{tabular}}
\end{table*}

Finally, we designed a series of experiments to validate the model's generalization and robustness. The generalization experiments primarily assessed the model's performance across varying node scales by comparing its results on the benchmark dataset CVRPLIB with known optimal solutions. In contrast, the robustness experiments involved perturbing the trained model by generating two additional distributions of node demands, while maintaining fixed node coordinate distributions and the ratio of node demands to vehicle capacities. This approach allowed us to evaluate the model's resilience to variations in demand distributions. 

As shown in Table \ref{vrplibresult}, on the one hand, the hypergraph model achieved a smaller average gap to the optimal solution compared to other learning-based state-of-the-art methods on various scales of the CVRPLIB dataset, indicating its strong generalization capability. The average gap of the proposed model is 5.01\%, compared with 7.51\% for E-GAT and 5.30\% for POMO with 8-fold augmentation. This result is important because the comparison includes instances whose scale and distribution differ from the random training distribution, suggesting that constraint-oriented hyperedges provide a useful inductive bias beyond the synthetic test set. On the uniformly generated CVRP100 benchmark in Table~\ref{basicresult}, our augmented model is also competitive: the objective is 15.77, only 0.04 higher than POMO with 8-fold augmentation. The remaining difference can be partly attributed to the uniform spatial distribution, where customer neighborhoods are less clearly separated, and the advantage of constraint-oriented hyperedge construction is weakened.  The absence of hyperedge embedding formation hinders the master node's representation learning, preventing it from adequately capturing the information with associated nodes. Consequently, the greedy inference strategy employed during decoding may overlook these nodes, leading to the emergence of suboptimal subpaths. To further examine generality and scalability under non-uniform geometry, we added an XML100 subset analysis~\cite{queiroga2022}, as reported in Table~\ref{tab:xml100_scalability}. XML100 keeps the node scale fixed at 100 but changes the geometric and demand distributions, allowing us to separate scale transfer from distributional geometry. On clustered customer layouts, the gap is reduced from 10.59\% with POMO to 10.11\% with the proposed model, and from 6.62\% to 6.48\% under 8-fold augmentation. On mixed random-clustered layouts, the gap is reduced from 8.75\% to 8.41\%, and from 5.77\% to 5.55\% under 8-fold augmentation. The corresponding standard deviations are also lower.  These results indicate that the hypergraph encoder is more effective when the customer distribution contains geometric clusters or mixed local structures. In such cases, pairwise dense attention tends to assign similar representations to nearby customers, which can lead to diffuse attention logits and over-smoothed node embeddings. Constraint-oriented hyperedges mitigate this effect by preserving group-level compatibility information and restricting representation aggregation to route-relevant customer sets.

These results also clarify the difference between heavy-encoder and heavy-decoder scalability. Heavy-decoder methods such as LEHD are designed to improve large-scale generalization by dynamically rebuilding relations during decoding, and they are particularly valuable when the test size is far larger than the training size. In contrast, the proposed framework is a heavy-encoder representation-learning approach: its advantage lies in learning constraint-aware node and hyperedge representations before decoding. When a fixed-scale encoder trained on 100-node synthetic instances is transferred to much larger or denser instances, the learned neighborhood and hyperedge distributions inevitably experience a distribution shift, which explains the performance decline on the largest CVRPLIB instance. Therefore, large-scale deployment should combine the proposed hypergraph representation with cross-scale training, hierarchical hypergraph construction, stronger decoding/search, or progressive adaptation strategies~\cite{luo2023lehd,zhou2023omni,liu2026scalenet,zhu2026evoreal}.

\begin{figure}[!ht]
    \centering
    \includegraphics[width=0.9\linewidth]{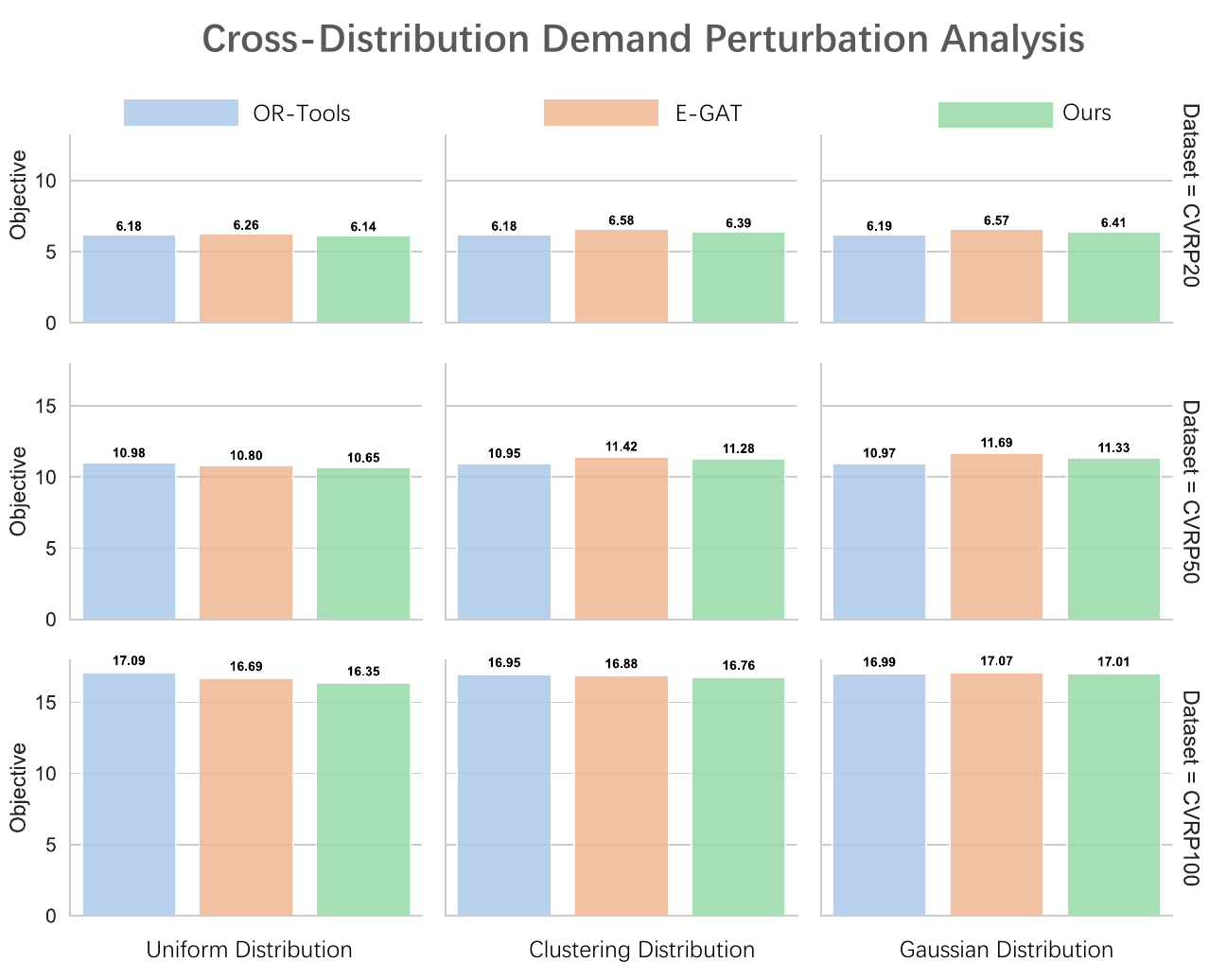}
    \caption{Cross-demand Perturbation Analysis}
    \label{fig:crossdemand}
\end{figure}

On the other hand, Fig.~\ref{fig:crossdemand} presents the model's robustness under various customer demand distributions. Notably, the Gaussian distribution of demands is characterized by a mean of 5.5 and a standard deviation of 1.33, while the cluster distribution centers are set at (3,5,7). Each cluster center serves as the mean for a normal distribution with a standard deviation of 0.1, ensuring that all demand values are clipped within the range of [1, 9]. This methodology aims to maintain cross-distribution noise interference in the ratio of node demands to vehicle capacities, thereby aligning more closely with real-world scenarios. If the demand range is designed to be too narrow, the CVRP would degenerate into a Traveling Salesman Problem (TSP). Conversely, an excessively broad demand range may result in vehicles returning to the depot after visiting only a few nodes, thereby compromising the validity of the capacity constraints.  

Experiments indicate that our method outperforms E-GAT even in the presence of demand-distribution perturbations. The hypergraph model exhibits a certain degree of robustness against interference, with performance degradation ranging from $3.42\%$ to $6.38\%$ on the noisy datasets. In comparison, the performance drop of E-GAT varied between $2.27\%$ and $8.24\%$, indicating less stable behavior under the same perturbation protocol. For reference, OR-Tools uses heuristic construction and Guided Local Search, making it less sensitive to such noise because the solution can be iteratively repaired. On small- and medium-scale datasets, the solution gap between our approach and OR-Tools ranges from $0.12\%$ to $3.55\%$, while E-GAT exhibits a gap of $0.47\%$ to $6.56\%$. These results suggest that constraint-oriented hyperedges provide a useful robustness bias by grouping customers according to capacity-related compatibility rather than relying only on pairwise spatial attention. However, autoregressive neural solvers still lack the iterative repair capability of improvement heuristics.
\subsubsection{Ablation Study}
\begin{table}[ht!]
\caption{Ablation Study}
    \label{tab:ablation}
\centering
    \begin{tabular}{l c c}
        \toprule
        Model  & CVRP 20 Gap(\%) & CVRP 50 Gap(\%) \\
        \midrule
        \textbf{Ours}     & 0.65 &   2.60             \\
            w/o dynamic hypergraph      &  2.95     &     7.42      \\
        w/o data augmentation  &   1.80       &   2.79    \\
        w/o dual-pointer recorder & 2.79       &     6.64    \\
        \bottomrule
    \end{tabular}
\end{table}
We conducted ablation studies on three key submodules of the hypergraph model by creating three variants and evaluating their effectiveness on randomly generated datasets in the greedy rollout policy. The results are summarized in Table \ref{tab:ablation} and variants presented as follows:
\begin{figure}[!ht]
    \centering
    \includegraphics[width=0.99\linewidth]{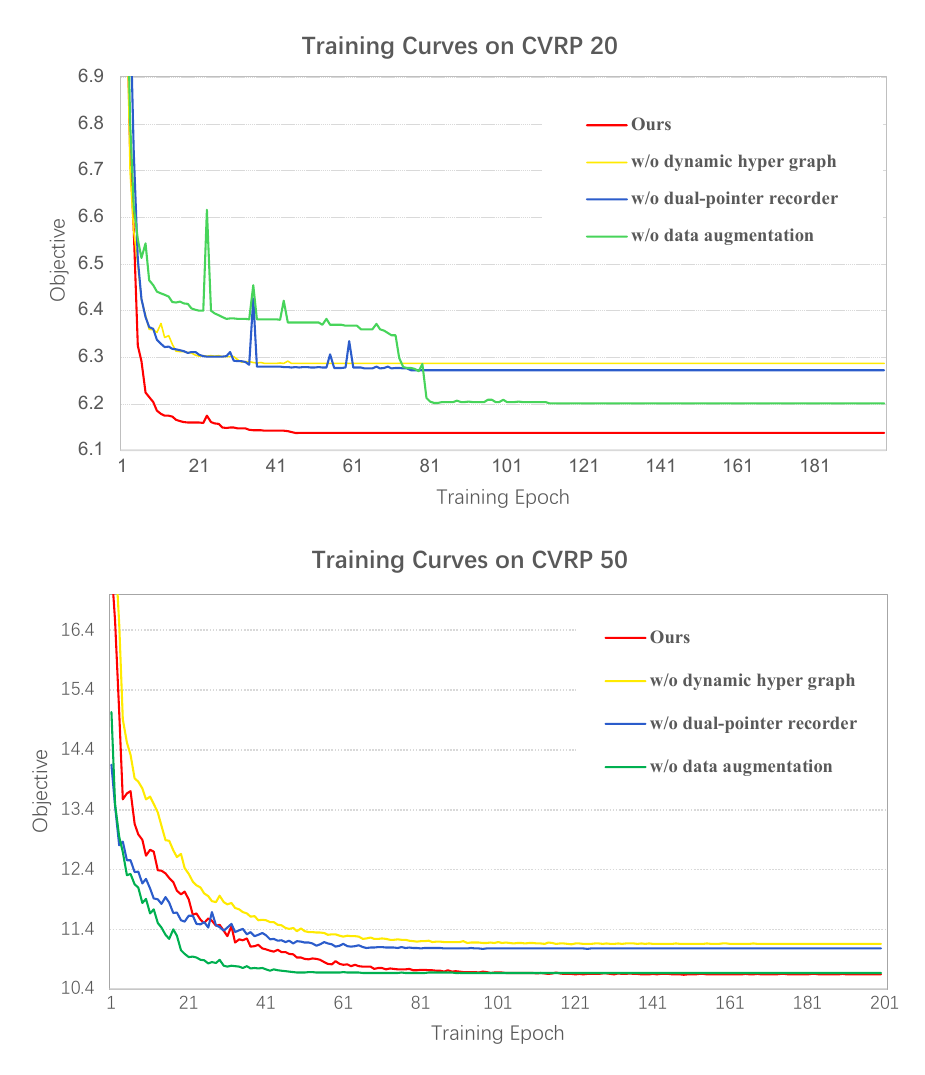}
    \caption{Training Curves of Various Size Datasets for Ablation}
    \label{fig:traincurve}
\end{figure}
\begin{itemize}
    \item The first variant, denoted as w/o dynamic hypergraph, replaces the hypergraph module with a standard GAT, thereby removing the dynamic hypergraph structure.
    \item The second variant, denoted as w/o data augmentation, eliminates data augmentation and uses only the raw node coordinates and demands as input features for the encoder.
    \item The third variant, denoted as w/o dual-pointer recorder, removes the route recorder in the decoder's dual-pointer mechanism. In this variant, decoding relies solely on a single pointer to compute attention based on the current node representation, without incorporating information from the partial solution.
\end{itemize}
Fig.~\ref{fig:traincurve} illustrates the training curves of the hypergraph model and its variants under the ablation study. The results indicate that the three innovative modules we proposed contribute positively to the overall architecture. The most direct evidence for the representation contribution is the removal of the dynamic hypergraph module: the gap increases from 0.65\% to 2.95\% on CVRP20 and from 2.60\% to 7.42\% on CVRP50. This degradation is larger than that caused by removing data augmentation on CVRP50, indicating that the improvement is not merely a consequence of extra input features or parameter count. Rather, preserving constraint-oriented hyperedge identities changes the information delivered to the decoder. Removing the dual-pointer recorder also causes a large degradation, from 2.60\% to 6.64\% on CVRP50, which confirms that route-history context is necessary for using high-order representations in sequential construction. However, the benefits of the data augmentation module are not pronounced in small-scale datasets, and it results in a slower convergence rate. We attribute this phenomenon to the limited action space in small-scale problems, where each decision step has a restricted number of actions, making action selection highly sensitive. The additional features introduced by the data augmentation module affect the attention scores, thereby increasing the number of action nodes in the decision space. This leads to instability during random sampling decisions. As the problem scale increases, the impact of this instability diminishes.
\subsubsection{Sensitivity Analysis}
In this section, we analyze the impact of hyperparameter tuning on the performance of the proposed model. Two key hyperparameters are introduced during model construction: \(\delta\) and \(\lambda\).  
\begin{figure}[!ht]
    \centering
    \includegraphics[width=1.0\linewidth]{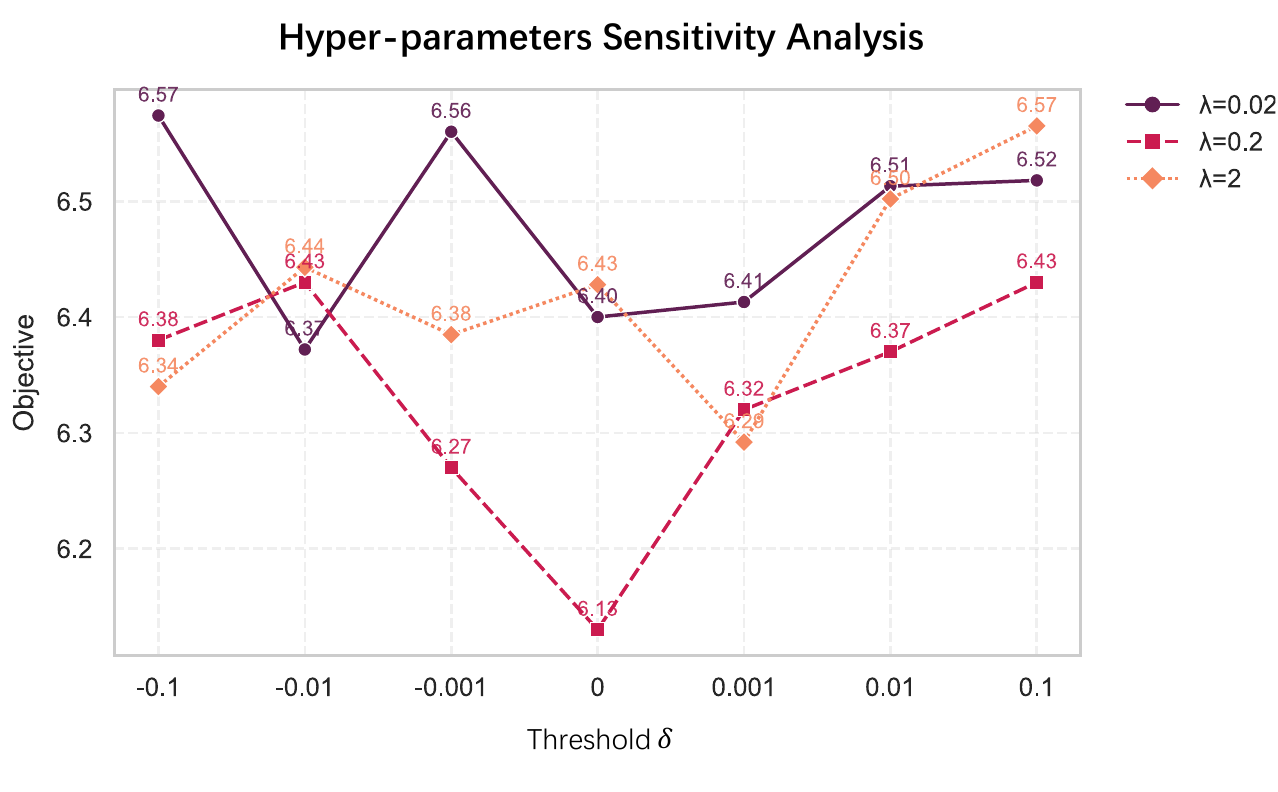}
    \caption{Hyperparameter Sensitivity Analysis}
    \label{fig:sensitive}
\end{figure}
\begin{itemize}
    \item \(\delta\): This parameter adjusts the weight threshold for reconstructing hyperedge nodes. Only nodes with weights exceeding the threshold \(\delta\) are included in the hyperedge. A higher \(\delta\) value results in fewer nodes being incorporated into the hyperedge, thereby tightening the constraints.  
    \item\(\lambda\): This coefficient balances the contributions of the $L1$ and $L2$ norms of the weight vector during hyperedge reconstruction, as defined in Equation (5). The value of \(\lambda\) determines the relative influence of sparsity ($L1$ norm) and smoothness ($L2$ norm) on the hyperedge reconstruction loss.  
\end{itemize}
The effects of these hyperparameters on model performance are demonstrated in Fig.~\ref{fig:sensitive}.  A grid search for \(\delta\) in the range \([-0.1, 0.1]\), combined with three \(\lambda\) configurations, revealed optimal performance at \(\delta = 0\) and \(\lambda = 0.2\). It is important to note that \(\delta\) represents the weight threshold for the adaptive hyperedge reconstruction, while the mapping for hyperedge reconstruction is composed of learnable weight parameters \(\Theta_{\mathrm{master}}\). We initialized these parameters using the Xavier uniform distribution, and during the weight learning process, we monitored the weight variation range, which remained between -0.1 and 0.1. Consequently, we conducted a hyperparameter sensitivity experiment on \(\delta\) using a magnitude-based grid search. The sensitivity settings for \(\lambda\) were informed by prior research~\cite{zhang2025learning}. Additionally, our model incorporates a weight hyperparameter \(\gamma = 0.01\), which scales the constraint penalties \(\mathcal{L}_{con}\) and hyperedge reconstruction error \(\mathcal{L}_{rec\_node}\) to the same order of magnitude, thereby preventing one type of loss from dominating the optimization of the hypergraph gradients.

The results indicate that an overly large threshold or excessive \(L_1\) regularization reduces the number of nodes per hyperedge, weakening encoder representation and restricting decoder decisions, thus degrading performance. Conversely, overly small thresholds and minimal \(L_1\) regularization introduce too many nodes into hyperedge reconstruction, causing over-smoothness in hypergraph representations. This sensitivity pattern is consistent with the motivation of the paper: useful hyperedges should be neither too sparse to miss feasible customer groups nor too dense to collapse into ordinary pairwise fully connected attention. The hyperedge threshold therefore controls a representation trade-off between constraint selectivity and information preservation.

\section{Conclusion}
This paper presents an end-to-end framework based on an encoder-decoder architecture to solve routing problems. In the encoder, a constraint-oriented hyperedge is constructed for each node to perform data augmentation and enhance node representation learning, which is further optimized through a hypergraph reconstruction loss. In the decoder, a dual-pointer mechanism is introduced to decode multi-head attention scores by simultaneously incorporating the partial solution state stored in the route recorder and the current node state. These scores are subsequently masked and fused through weighted aggregation, enabling the progressive construction of routing solutions. 
The main significance of the work is to introduce a constraint-aware high-order representation perspective into neural routing. Hypergraph learning structures the input representation by preserving constraint-induced customer groups, whereas reinforcement learning optimizes the sequential route-construction policy through label-free route-level feedback. The results show clear benefits on small- and medium-scale CVRP instances, strong improvements on CVRPTW, and an average CVRPLIB gap competitive with strong neural baselines. They also reveal limitations: large-scale transfer remains challenging when a fixed-scale hypergraph encoder is combined with greedy decoding. Future work will therefore focus on hierarchical hypergraph construction and stronger decoding/search strategies.


\bibliographystyle{elsarticle-num} 
\bibliography{ref}





\end{document}